\documentclass[letterpaper, 10 pt]{ieeeconf}  

\IEEEoverridecommandlockouts                              

\usepackage{amsmath,amsfonts,amssymb}
\usepackage{algorithmic,algorithm}
\usepackage[export]{adjustbox} 

\usepackage{lineno}
\usepackage{soul,color}

\usepackage[dvipsnames]{xcolor}
    
\usepackage{multirow, hhline}
\usepackage{subfloat}
\usepackage{mathtools, mathdots}

\usepackage{caption}
\usepackage{subcaption}
\usepackage{textcomp}
\usepackage{gensymb}
\usepackage{diagbox}
\usepackage{siunitx}
\usepackage{lettrine}

\newcommand\mydots{\makebox[1em][c]{.\hfil.\hfil.}}

\makeatletter
\let\NAT@parse\undefined
\makeatother
\usepackage{hyperref}
\hypersetup{
    colorlinks=true,
    linkcolor=blue,
    filecolor=magenta,
    urlcolor=blue,
}

\begin{document}

\title{A General Relationship between Optimality Criteria and Connectivity Indices for Active Graph-SLAM}

\author{
    Julio~A.~Placed,~\IEEEmembership{Student Member,~IEEE,}
    and~Jos\'e~A.~Castellanos,~\IEEEmembership{Senior Member,~IEEE}
    \thanks{The authors are with the Instituto de Investigaci\'on en Ingenier\'ia de Arag\'on (I3A), Universidad de Zaragoza, C/Mar\'ia de Luna 1, 50018, Zaragoza, Spain e-mail: \{jplaced,jacaste\}@unizar.es.}
}

\markboth{}%
{}

\maketitle

\begin{abstract}
    Quantifying uncertainty is a key stage in active simultaneous localization and mapping (SLAM), as it allows to identify the most informative actions to execute. However, dealing with full covariance or even Fisher information matrices (FIMs) is computationally heavy and easily becomes intractable for online systems.
    In this work, we study the paradigm of active graph-SLAM formulated over \textit{SE(n)}, and propose a general relationship between the FIM of the system and the Laplacian matrix of the underlying pose-graph.
    This link makes possible to use graph connectivity indices as utility functions with optimality guarantees, since they approximate the well-known optimality criteria that stem from optimal design theory.
    Experimental validation demonstrates that the proposed method leads to equivalent decisions for active SLAM in a fraction of the time.
\end{abstract}


\section{Introduction} \label{S:1}

\lettrine{A}{utonomous} exploration of unknown environments has long attracted the attention of the robotics community and despite being essential for achieving high-level autonomy in numerous applications (e.g., search and rescue), it still remains an open problem. Mainly, it involves planning and navigating under uncertainty, thus making necessary to have a model of the surrounding environment.

Simultaneous Localization and Mapping (SLAM) passively deals with the base problem of incrementally building a map of the environment while at the same time locating the robot on it. Many approaches have been developed to solve it, although the interest of this paper particularly lies on (pose) graph-SLAM methods, that intuitively formulate the problem using a graph representation where nodes encode the robot poses, and edges encode the constraints between them. This SLAM variant is built on the idea that the map representation can be retrieved once the robot states have been properly estimated~\cite{grisetti10}.
After solving the data association and building the graph, it all comes down to finding the optimal nodal configuration via maximum likelihood (ML), i.e., finding the set of robot poses that minimizes a cost function of the observations. See~\cite{grisetti10, durrant06, cadena16} and the references there in.

Active SLAM expands the previous problem so as to include the selection of the actions the robot should execute to build the best model of the environment possible. This new decision-making problem, erected on \textit{active perception}~\cite{bajcsy88}, can be formally defined as the control of a robot that is performing SLAM with the objective of reducing the uncertainty of its location and of the map representation. Given the probabilistic nature of SLAM, the above involves reasoning over probabilistic states (i.e., \emph{beliefs}) under uncertainty, a problem also referred to in the literature as belief space planning (BSP)~\cite{platt10}.
Active SLAM has been traditionally
divided in three phases for the ease of its resolution~\cite{makarenko02}:
\begin{enumerate}
    \renewcommand\labelenumi{\roman{enumi})}
    \item the identification of possible locations to explore,
    \item the evaluation of the utility associated to the actions that would take the robot from its current position to each of those locations, and
    \item the selection and execution of the optimal actions.
\end{enumerate}

In the first stage, actions to reach all possible locations should be evaluated, although easily proves to be intractable due to the dimensionality of the state and action spaces.
In practice, the most common approach is the identification of the points which lie between the known and unknown regions of the map, i.e., frontiers.
During the second step, the utility of each set of actions is estimated by quantifying the expected uncertainty in the two target random variables: the robot location and the map. Uncertainty quantification can be based either on theory of optimal experimental design (TOED)~\cite{chen20} or information theory (IT)~\cite{stachniss05}, although it is the former on which this letter focuses.
Either way, these utility functions must find the equilibrium between exploring new areas and exploiting those previously seen (i.e., the so called \emph{exploration-exploitation dilemma}).
Finally, third phase consists in executing the set of actions with highest utility.

TOED-based utility functions, or \emph{optimality criteria}
directly map
the expected covariance matrices to the real scalar space through their eigenvalues.
However, the use (i.e., propagation, store and analysis) of these dense matrices quickly becomes intractable in online active SLAM. For example, approaches that require computing the determinant of the \textit{a posteriori} covariance matrix are $\mathcal{O}(n^3)$ complex in general, with $n$ the dimension of the full state. In an effort to lessen the computational load, most works resort to the sparser Fisher information matrices (FIMs) \cite{indelman15} or use sparsified representations~\cite{carrillo18, elimelech19}. Even so, the use of TOED metrics is costly and thus often disregarded in the literature in favor of fast IT metrics, e.g., the entropy.

In this work, we demonstrate that the expensive evaluation of optimality criteria
over the expected FIMs during active graph-SLAM can be approximated by analyzing the topology of the expected pose-graphs. This method requires way less resources and eases the use of optimality criteria in online methods.
First, we propose a general relationship between the FIM of a graph-SLAM system and the Laplacian matrix of the underlying pose-graph. This allows to relate their spectra under certain conditions and, therefore, to establish a link between optimality criteria and graph connectivity indices.


The rest of the paper is organized as follows. Section~\ref{S:2} discusses related work and the contributions of this paper, Section~\ref{S:3} presents preliminary contents, Section~\ref{S:4} details the theoretical contributions of this paper, and Section~\ref{S:5} contains the experimental results. The manuscript is concluded in Section~\ref{S:6}, where future work is also outlined.

\section{Related Work} \label{S:2}

The idea of the topology of a graph being closely related to its optimality was already noticed four decades ago, when Cheng~\cite{cheng81} realized that a graph with the maximum number of spanning trees is generally optimal, and thus related two problems (from graph theory and TOED) that had always been viewed differently. More recently, Khosoussi \textit{et al.}~\cite{khosoussi14} observed that certain classical optimality criteria
are closely related to the connectivity of the underlying pose-graph. In particular, they show the existing relationship between the number of spanning trees of the pose-graph and the determinant of the covariance matrix of the SLAM system (traditionally known as \textit{D}-optimality), and that between its algebraic connectivity and the covariance's maximum eigenvalue; for 2D and assuming constant and isotropic variance through measurements.
In~\cite{khosoussi19}, the authors extend the above relationships to the case in which uncertainty evolves as the trajectory does. Given block-isotropic Gaussian noise in the measurements, they relate the determinant of the covariance to the determinants of the Laplacians of two pose-graphs, each weighted by the decoupled rotational/translational inverse variance.
Chen \textit{et al.}~\cite{chen21} study graph-SLAM as the synchronization problem over $\mathbb{R}^n\times SO(n)$. They propose an approximation relationship for the FIM's trace (\textit{T}-optimality) and two bounds on its determinant that, once again, depend on the Laplacians of two weighted pose-graphs. In contrast to~\cite{khosoussi19}, they assume isotropic Langevin noise for orientation and block-isotropic Gaussian noise for translation.

Fast exploitation of the graph structure was also recently transferred to the domain of information-theoretic BSP. Kitanov and Indelman~\cite{kitanov19} prove that the number of spanning trees is also a good approximation of the posterior entropy ---a reasonable fact since it ultimately depends on the covariance determinant as classical \textit{D}-optimality. They also present a relationship between the graph's node degree and the Von Neumann entropy. Despite being faster, it fails to select the optimal actions in some cases (just as \textit{T}-optimality would).

Following all the above relationships, instead of maximizing optimality criteria of the FIM, the optimal set of actions in active SLAM can be found more efficiently through maximizing graph connectivity indices. 
Chen \textit{et al.}~\cite{chen20} build a 2D multi-robot active SLAM algorithm that achieves uncertainty minimization and information sharing between agents thanks to the fast evaluation of the underlying pose-graphs topology.

All the aforementioned works, however, have isolatedly related specific classic optimality criteria to certain connectivity indices; also for specific/restrictive SLAM configurations. In this letter, on the basis of graph theory, differential models,~\cite{khosoussi14} and~\cite{placed21}, we derive a general theoretical relationship between the FIM of a graph-SLAM problem formulated over the Lie group \textit{SE(n)}, and the Laplacian
of the underlying pose-graph.
On top of that, we establish a strong link between the spectrum of both matrices and relate optimality criteria
to graph connectivity indices.
Contrarily to previous works, measurement noises are not restricted to be \mbox{(block-)} isotropic nor constant, formulation is done over \textit{SE(n)}, and modern optimality criteria are used. Note that these differences are key for active SLAM applications, since: (i) covariance is usually non-isotropic (i.e., variances may not be the same in all directions for rotation/translation and may be cross-correlated) and varies along exploration, (ii) only differential representations maintain the monotonicity of the decision making criteria (see~\cite{rodriguez18}), and (iii) the use of traditional criteria is not suitable for active SLAM\footnote{The best known example is the possibility of a single element driving classical \textit{D}-optimality to zero. Also, the size of the FIM grows over time, so comparison of raw determinants
is unfair~\cite{mihaylova02}.}.
We validate the proposed relationships and analyze time complexity in several 2D and 3D SLAM datasets.
On average, our method requires just $10\%$ of the time traditional computations would, and error is only $2\%$. Besides, graph-based approximations always maintain same the trend of optimality criteria over time, which makes their use appropriate for active SLAM.

\section{Preliminaries}\label{S:3}

\subsection{Graph-based SLAM}\label{SS:3a}

Graph-SLAM methods employ a graph representation to solve the SLAM estimation problem. Nodes in the graph represent the poses of the variables of interest (i.e., the robot location and the map points), while edges represent the sensor measurements. The process of generating constraints between nodes from the observations is known as data association and is usually bounded to the most likely topology to restrain complexity. Thus, under the assumption of observations affected by Gaussian noise, an edge will encode the relative pose between two nodes and the associated covariance matrix (or FIM). Such constraints may be related to a sensor measurement (e.g., odometry) or to a loop closure. The work in this manuscript is particularly focused on pose-graphs, which flatten the above representation encoding only robot poses in vertices. These sparser representations can be achieved by marginalizing the map points in landmark-based representations, or by building a discretized metric map via scan matching and updating it after a loop closure occurs.

In any case, once the graph is built and the data association problem solved, the goal
is to compute the (Gaussian-approximated) posteriors over the robot poses. That is, to find the nodal configuration that maximizes the likelihood of the observations~\cite{grisetti10}. The following optimization problem may be solved using, e.g., Gauss-Newton method:

\begin{gather}
    \boldsymbol{x}^* = \arg\min_{\boldsymbol{x}} \boldsymbol{F}(\boldsymbol{x}) \label{eq:mlopt}\\
    \text{s.t.} \quad \boldsymbol{F}(\boldsymbol{x}) = \frac{1}{2} \textstyle\sum\limits_{j=1}^m \boldsymbol{F}_j(\boldsymbol{x}) \nonumber = \displaystyle\frac{1}{2}\textstyle\sum\limits_{j=1}^m\boldsymbol{e}_j^T(\boldsymbol{x}) \boldsymbol{\Sigma}_j^{-1} \boldsymbol{e}_j(\boldsymbol{x})
\end{gather}
\noindent where $\boldsymbol{x}$ are the variables of interest (i.e., the robot poses), $\boldsymbol{F}$ the cost function of the $m$ observations, and $\boldsymbol{e}_j$ and $\boldsymbol{\Sigma}_j$ the error to be minimized and covariance for each measurement.

\subsection{Modern Optimality Criteria in Active SLAM}\label{SS:3b}

While performing active SLAM, decision making comes down to computing the utility that executing a certain set of actions would lead to, i.e., to quantifying their expected uncertainty. Kiefer~\cite{kiefer74}, on the basis of TOED, shows that there is a family of mappings that quantify uncertainty, i.e., $\|\boldsymbol{\Sigma}\|\to\mathbb{R}$, which are dependent of just one parameter ($p$):

\begin{equation}
    \|\boldsymbol{\Sigma}\|_p \triangleq \left( \frac{1}{\ell}\text{trace}(\boldsymbol{\Sigma}^p) \right) ^ \frac{1}{p}
\end{equation}
\noindent where $\ell$ is the dimension of the state vector and $\boldsymbol{\Sigma}\in\mathbb{R}^{\ell\times\ell}$ the covariance matrix which measures its uncertainty. The preferred set of actions will be that with lowest $\|\boldsymbol{\Sigma}\|_p$.

Kiefer's information function may be expressed in terms of the eigenvalues of $\boldsymbol{\Sigma}$, $(\lambda_1,\mydots,\lambda_\ell)$, and particularized for the different values of $p$; yielding four modern optimality criteria~\cite{pukelsheim06}:
\begin{itemize}
  \item \textit{T}-optimality criterion ($p=1$): captures the average variance. Its computation is fast,
  but a single element may drive the whole metric and thus it may perform similar to just evaluating the highest eigenvalue~\cite{carrillo12}.
  \begin{equation}
    T{\text -} opt \triangleq \frac{1}{\ell}\textstyle\sum\limits_{k=1}^\ell \lambda_k \label{eq:topt}
  \end{equation}

  \item \textit{D}-optimality criterion ($p=0$): captures the volume of the covariance hyper ellipsoid.
  Only
  it captures global uncertainty and holds the monotonicity property under both absolute and differential representations~\cite{rodriguez18}.
  \begin{equation}
    D{\text -} opt \triangleq \exp \left(\frac{1}{\ell} \textstyle\sum\limits_{k=1}^\ell \log(\lambda_k) \right) \label{eq:dopt}
  \end{equation}

  \item \textit{A}-optimality criterion ($p=-1$): captures the harmonic mean variance, being thus sensitive to smallest eigenvalues ---in contrast to $T\text{-}opt$ which just neglects them--- and insensitive to extremely large ones.
  \begin{equation}
    A{\text -} opt \triangleq \left(\frac{1}{\ell}\textstyle\sum\limits_{k=1}^\ell \lambda_k^{-1}\right)^{-1} \label{eq:aopt}
  \end{equation}

  \item $E$-optimality criterion ($p\to\pm\infty$): approximates the uncertainty using a single eigenvalue. Despite its computation is fast, this criterion tends to be too optimistic by underestimating the covariance (for the case of the minimum eigenvalue).
  \begin{align}
    E{\text -} opt \triangleq \min (\lambda_k : k=1,...,\ell) \\
    \tilde{E}{\text -} opt \triangleq \max (\lambda_k : k=1,...,\ell) \label{eq:eopt}
  \end{align}

\end{itemize}

\subsection{Graph Theory}\label{SS:3c}

A strict undirected graph is defined by the ordered pair of sets ${\mathcal{G}}\triangleq(\mathcal{V},\mathcal{E})$, where $\mathcal{V}=\{\boldsymbol{v}_0,...,\boldsymbol{v}_n\}$ is the set of vertices and ${\mathcal{E}=\{\boldsymbol{e}_1,...,\boldsymbol{e}_m\}\subset\{\ \{\boldsymbol{v}_i,\boldsymbol{v}_k\} \ |\ \boldsymbol{v}_i,\boldsymbol{v}_k\in\mathcal{V},\boldsymbol{v}_i\neq \boldsymbol{v}_k\}}$ the set of edges. Their dimensions will be $|\mathcal{V}|=n$ and $|\mathcal{E}|=m$.
The adjacency matrix of the graph, $\textbf{A}\in\{0,1\}^{n\times n}$, indicate whether pairs of vertices are connected or not. Each element $a_{ik}$ will be $1$ if the pair $(\boldsymbol{v}_i, \boldsymbol{v}_k)$ is connected and $0$ otherwise. Note that the diagonal will be zero.
The incidence matrix, $\boldsymbol{Q}$, shows the connections between vertices and edges and can be defined as a concatenation of $m$ column vectors, each of them associated to an edge:
\begin{equation}
    \boldsymbol{Q}=[\boldsymbol{q}_1, \boldsymbol{q}_2, \mydots, \boldsymbol{q}_m]\in\{-1,0,1\}^{n\times m}
\end{equation}

\noindent The column block associated to the edge $\boldsymbol{e}_{ik}\equiv\boldsymbol{e}_j$, that connects $\boldsymbol{v}_i$ and $\boldsymbol{v}_k$, will be denoted as $\boldsymbol{q}_j$. All elements of $\boldsymbol{q}_j$ will be zero except those associated to the vertices incident upon $\boldsymbol{e}_j$ (i.e., the $i$-th and $k$-th) which will be \mbox{$[q_j]_i=-[q_j]_k=1$}. The Laplacian matrix of ${\mathcal{G}}$ is a matrix representation of the whole graph, and may be read as a particular case of the discrete Laplace operator. It can be expressed in terms of $\boldsymbol{Q}$ as:
\begin{equation}
    \boldsymbol{L}\triangleq \boldsymbol{Q} \boldsymbol{Q}^T = \boldsymbol{q}_1 \boldsymbol{q}_1^T + \boldsymbol{q}_2 \boldsymbol{q}_2^T + ... + \boldsymbol{q}_m \boldsymbol{q}_m^T \in\mathbb{Z}^{n\times n}
    \label{eq:laplacian1}
\end{equation}
Or, more compactly, as:
\begin{equation}
    \boldsymbol{L} \triangleq \textstyle\sum\limits_{j=1}^m \boldsymbol{E}_j = \textstyle\sum\limits_{j=1}^m \boldsymbol{q}_j \boldsymbol{q}_j^T \label{eq:laplacian}
\end{equation}
where each generator $\boldsymbol{E}_j\in\{-1,0,1\}^{n\times n}$ represents the connection between the pair $(\boldsymbol{v}_i,\boldsymbol{v}_k)$ through the edge $\boldsymbol{e}_j$. An element of the matrix diagonal will be $1$ if it is associated to the vertices, i.e., $[E_j]_{i,i}$ and $[E_j]_{k,k}$; and $0$ otherwise. Off-diagonal elements will be $-1$ if the nodes are related, i.e., $[E_j]_{ik}$ and $[E_j]_{k,i}$; and $0$ otherwise.

For a weighted graph ${\mathcal{G}}_\gamma$ in which $\tilde{\boldsymbol{e}}_j\triangleq(\boldsymbol{v}_i,\boldsymbol{v}_k,\gamma_{ik})$ with $\gamma_j\equiv \gamma_{ik} \in\mathbb{R}$, generalization is straight-forward. The weighted Laplacian will be now given by:
\begin{equation}
    \boldsymbol{L}_\gamma \triangleq \sum_{j=1}^m \boldsymbol{E}_j \gamma_j = \left\{
        \begin{array}{ll}
            -\gamma_{ik} & \text{if} \ i\neq k, a_{ik}=1 \\
            0 & \text{if} \ i\neq k,  a_{ik}=0 \\
            \textstyle\sum_{q=1}^n \gamma_{iq} & \text{if} \ i=k
        \end{array} \right. \label{eq:wlaplacian}
\end{equation}
\noindent Note that~\eqref{eq:wlaplacian} yields to~\eqref{eq:laplacian} when $\gamma_j=1\ \forall j$. Also, $\boldsymbol{L}_\gamma$ is positive semi-definite and singular, since $\boldsymbol{L}_\gamma \boldsymbol{1}^T = \boldsymbol{0}^T$.

\subsection{Spectral Graph Theory} \label{SS:2d}

Most important graph connectivity indices come from the analysis of the Laplacian spectrum, since it reflects how a graph is connected. Consider $\boldsymbol{\mu} = (\mu_1, \mu_2, \mydots,\mu_n)$ the ordered set of eigenvalues of $\boldsymbol{L}$ and $\boldsymbol{\tilde{\mu}} = (\tilde{\mu}_1, \tilde{\mu}_2, \mydots,\tilde{\mu}_n)$ that of $\boldsymbol{L}_\gamma$; both ranked in increasing order. In connected graphs, the Laplacian matrix has one zero eigenvalue with unit eigenvector, i.e., $\mu_1 = \tilde{\mu}_1 = 0$.

The simplest metric broadly studied in the literature is the sum of the Laplacian eigenvalues~\cite{zhou08,ganie16}. The sum of all non-zero eigenvalues is known to be given by:
\begin{align}
    S = \textstyle\sum\limits_{k=2}^n \mu_k = \text{tr}(\boldsymbol{L}) = \sum_k l_{kk} = 2m
\end{align}
\noindent being $l_{kk}$ the diagonal elements of $\boldsymbol{L}$. Since the traces of $\boldsymbol{L}$ and $\boldsymbol{L}_\gamma$ are proportional, as shown hereafter, the previous metric can be easily generalized for a weighted graph.
\begin{align}
    \text{tr}(\boldsymbol{L}_\gamma) = \textstyle\sum\limits_{j=1}^n \sum\limits_{k=1}^n \gamma_j l_{kk}=\frac{1}{n}\sum\limits_{j=1}^n \gamma_j \sum\limits_{k=1}^n l_{kk}
    &\propto \text{tr}(\boldsymbol{L})
\end{align}

A second important index is the number of spanning trees, $t({\mathcal{G}})$, i.e., the number of sub-graphs that are also trees with minimum number of edges and which set of vertices equals that of the original graph. This index provides a measure for the global reliability of a network. Following Kirchhoff's matrix-tree theorem (MTT), it is given by the determinant of the reduced Laplacian matrix (after anchoring an arbitrary vertex), which equals to any cofactor of $\boldsymbol{L}$:
\begin{equation}
    t({\mathcal{G}}) \triangleq \det(\boldsymbol{L}_{reduced}) = \text{cof}(\boldsymbol{L}) =  \frac{1}{n}\textstyle\prod\limits_{k=2}^n \mu_k
\end{equation}
The weighted MTT allows to compute the weighted number of spanning trees as:
\begin{equation}
    t({\mathcal{G}}_\gamma) \triangleq \text{cof}(\boldsymbol{L}_\gamma) =  \frac{1}{n}\textstyle\prod\limits_{k=2}^n \tilde{\mu}_k
\end{equation}

The second smallest eigenvalue of the Laplacian is also a crucial index of a graph, since its value reflects whether it is disconnected~\cite{de07}. It is known as the algebraic connectivity and is greater than zero only for connected graphs:
\begin{equation}
    \alpha({\mathcal{G}}) \triangleq \min(\mu_k : k=2,\mydots,n)=\mu_2
\end{equation}
\noindent Its generalization for weighted graphs is straight-forward.

Finally, the Kirchhoff index, $K({\mathcal{G}})$, measures the resistance between each pair of vertices under the assumption that edges are unit resistors, and is defined by~\cite{chen07}:
\begin{equation}
    K({\mathcal{G}}) \triangleq n\textstyle\sum\limits_{k=2}^n \mu_k^{-1}
\end{equation}

\section{Linking the Fisher Information Matrix and the Graph Laplacian}\label{S:4}

Consider a typical SLAM pose-graph in which nodes encode robot poses and edges the relative transformation between node pairs and its uncertainty, usually in the form of a FIM.
The robot pose and its uncertainty can be defined with Lie groups using the special Euclidean group $SE(n')$.
As in differential representations, the real location of the robot w.r.t. a global frame ($w$), denoted as $\boldsymbol{T}_{wi}$, will be defined by a large noise-free value which contains the estimated location, and a small perturbation that encodes the estimation error:
\begin{equation}
    \boldsymbol{T}_{wi} = \bar{\boldsymbol{T}}_{wi} \exp{\left({\boldsymbol{d}_i}^\wedge\right)}
\end{equation}
\noindent where $\bar{\boldsymbol{T}}_{wi}\in SE(n')$ is a large mean transformation, and $\boldsymbol{d}_i\in\mathbb{R}^\ell$ is a random vector normally distributed and defined by its mean $\bar{\boldsymbol{d}}_i$
and covariance ${\boldsymbol{\Sigma}_i=\mathbb{E}[(\boldsymbol{d}_i-\bar{\boldsymbol{d}}_i)(\boldsymbol{d}_i-\bar{\boldsymbol{d}}_i)^T]}$; expressed in its own frame.
The hat operator $(\cdot)^\wedge$ defines the mapping from the real vector space to that of the Lie algebra~\cite{sola18}.
Alternatively, the perturbation may be expressed in the global frame, just like in differential representations~\cite{barfoot14}, in which case the exponential map will precede the estimate:
\begin{equation}
    \boldsymbol{T}_{wi} =  \exp{\left({\boldsymbol{d}_{wi}}^\wedge\right)} \ \bar{\boldsymbol{T}}_{wi} \label{eq:se3}
\end{equation}

As mentioned previously, the $j$-th edge in the pose-graph will encode a single measurement that represents the relative pose change between the pair $(\boldsymbol{v}_i,\boldsymbol{v}_k)$. This transformation can be simply expressed as:
\begin{equation}
    \boldsymbol{T}_{ik} = \boldsymbol{T}_{wi}^{-1} \ \boldsymbol{T}_{wk}= \exp{\left( {\boldsymbol{d}_{ik}}^\wedge\right)}\bar{\boldsymbol{T}}_{ik} \label{eq:relativeT}
\end{equation}
assuming that measurements are Gaussian on $SE(n')$, i.e., \mbox{$\boldsymbol{d}_{ik}\sim\mathcal{N}(0,\boldsymbol{\Sigma}_{ik}')$} with $\boldsymbol{\Sigma}_j'\equiv\boldsymbol{\Sigma}_{ik}'\in\mathbb{R}^{\ell\times \ell}$ the measurement covariance matrix. Note $\boldsymbol{d}_{ik}$ will be referenced to the $i$-th frame, as the mean transformation is perturbed on the left.

The error term of each measurement can be defined from~\eqref{eq:relativeT} as the difference between that single measurement, $\bar{\boldsymbol{T}}_{ik}$, and the optimal estimate, $\boldsymbol{T}_{ik}$:
\begin{equation}
    \boldsymbol{e}_{j}(\boldsymbol{x}) \equiv \boldsymbol{e}_{ik}(\boldsymbol{x}) = \ln\left(  \boldsymbol{T}_{wi}^{-1}\boldsymbol{T}_{wk}{\bar{\boldsymbol{T}}_{ik}}^{-1}\right)^\vee
\end{equation}
with $\boldsymbol{x}=\left(\boldsymbol{T}_{w1},\mydots, \boldsymbol{T}_{wn}\right)$ the variables of interest, and $(\cdot)^\vee$ the inverse of the hat operator. Inserting~\eqref{eq:se3} into the above,
\begin{equation}
    \boldsymbol{e}_{j}(\boldsymbol{x}) = \ln\left(
    \bar{\boldsymbol{T}}_{wi}^{-1} \exp\left(-{\boldsymbol{d}_{wi}}^\wedge\right)
    \exp\left({\boldsymbol{d}_{wk}}^\wedge\right)\bar{\boldsymbol{T}}_{wk}{\bar{\boldsymbol{T}}_{ik}}^{-1}\right)^\vee \label{eq:ej1}
\end{equation}
and using now the adjoint (see, e.g.,~\cite{sola18}) to transform vectors from the tangent space around one element to that of another, and the
definition $\exp{(Ad_{\boldsymbol{A}}  \boldsymbol{B})} \triangleq \boldsymbol{A}  \exp{(\boldsymbol{B})} \boldsymbol{A}^{-1}$, the term inside the logarithm in~\eqref{eq:ej1} becomes:
\begin{align}
    \exp\left(-Ad_{\boldsymbol{T}_{wi}^{-1}}{\boldsymbol{d}_{wi}}^\wedge\right)\exp\left(Ad_{\boldsymbol{T}_{wi}^{-1}}{\boldsymbol{d}_{wk}}^\wedge\right) \exp\left({\boldsymbol{e}}_j(\bar{\boldsymbol{x}}){}^\wedge\right)
\end{align}
with $\boldsymbol{e}_j(\bar{\boldsymbol{x}})=\ln( \bar{\boldsymbol{T}}_{wi}^{-1}\bar{\boldsymbol{T}}_{wk}{\bar{\boldsymbol{T}}_{ik}}^{-1})^\vee$ and $(\bar{\boldsymbol{T}}_{wi}^{-1}\bar{\boldsymbol{T}}_{wk}{\bar{\boldsymbol{T}}_{ik}}^{-1})$ small.

Finally, using the first-order approximation~\cite{brossard17} of the Baker-Campbell-Haussdorf formula for the product of exponential maps, the linearized error can be expressed as:
\begin{equation}
    \boldsymbol{e}_{j}(\boldsymbol{x}) \approx Ad_{\boldsymbol{T}_{wi}^{-1}}({\boldsymbol{d}_{wk}}-{\boldsymbol{d}_{wi}}) + {\boldsymbol{e}}_j(\bar{\boldsymbol{x}})
\end{equation}
Or, equivalently, in matrix form,
\begin{equation}
    \boldsymbol{e}_j(\boldsymbol{x}) \approx \boldsymbol{e}_j(\bar{\boldsymbol{x}}) - Ad_{\boldsymbol{T}_{wi}^{-1}} \begin{bmatrix} \pmb{\mathbb{I}}_{\ell} & -\pmb{\mathbb{I}}_{\ell} \end{bmatrix} \delta \boldsymbol{x}_j \label{eq:errormatrixform}
\end{equation}
with $\delta \boldsymbol{x}_j = [ \boldsymbol{d}_{wi} \ \boldsymbol{d}_{wk} ]^T$ and $\pmb{\mathbb{I}}_{\ell}$ the identity matrix of size $\ell$.

Inserting the error function into~\eqref{eq:mlopt} and generalizing $\delta\boldsymbol{x}_j$ to $\delta\boldsymbol{x}=\left[ \boldsymbol{d}_{w1}\ \mydots \ \boldsymbol{d}_{wn}\right]^T$ to account for all measurements, the ML (quadratic) cost function will be:
\begin{align}
    \boldsymbol{F}(\boldsymbol{x}) &\approx \boldsymbol{F}(\bar{\boldsymbol{x}}) - \textstyle\sum_j \boldsymbol{e}_j(\bar{\boldsymbol{x}})^T {\boldsymbol{\Sigma}_j'}^{-1} Ad_{\boldsymbol{T}_{wi}^{-1}} \boldsymbol{\mathcal{I}}_j \delta\boldsymbol{x} \nonumber \label{eq:costF}\\
    & \quad+ \frac{1}{2} \textstyle\sum_j \delta\boldsymbol{x}^T {\boldsymbol{\mathcal{I}}_j}^T Ad_{\boldsymbol{T}_{wi}^{-1}}^T {\boldsymbol{\Sigma}_j'}^{-1}Ad_{\boldsymbol{T}_{wi}^{-1}}\boldsymbol{\mathcal{I}}_j\delta\boldsymbol{x} \\
    &= \boldsymbol{F}(\bar{\boldsymbol{x}}) - \boldsymbol{Z}\delta\boldsymbol{x} + \frac{1}{2}\delta\boldsymbol{x}^T\boldsymbol{Y}\delta\boldsymbol{x}
\end{align}
with $\boldsymbol{\mathcal{I}}_j$ the $1\text{-by-}n$ selection block matrix, populated with zero blocks everywhere but in the $i$-th and $k$-th columns, where $[\boldsymbol{\mathcal{I}}_j]_{1,i} = -[\boldsymbol{\mathcal{I}}_j]_{1,k} = \pmb{\mathbb{I}}_{\ell}$.

The Fisher information matrix ---or Hessian--- of the entire system can be directly extracted from~\eqref{eq:costF} and expressed as:
\begin{equation}
    \boldsymbol{Y} = \textstyle\sum\limits_{j=1}^m \boldsymbol{Y}_j = \textstyle\sum\limits_{j=1}^m {\boldsymbol{\mathcal{I}}_j}^T {\boldsymbol{\Sigma}_j}^{-1}\boldsymbol{\mathcal{I}}_j
\end{equation}
\noindent with {${\boldsymbol{\Sigma}_j}^{-1}=Ad_{\boldsymbol{T}_{wi}^{-1}}^T {\boldsymbol{\Sigma}_j'}^{-1}Ad_{\boldsymbol{T}_{wi}^{-1}}$} the inverse covariance matrix of the relative movement, expressed in $w$.
Since we kept the perturbations $\delta\boldsymbol{x}$ in the global frame from~\eqref{eq:errormatrixform} on, the need arises to express their covariance in that frame as well.
This formulation over Lie groups is analogous to the differential one (see, e.g.,~\cite{grisetti10}), but embedding the equivalent measurement Jacobian in the covariance rather than in $\boldsymbol{\mathcal{I}}_j$.

\begin{figure*}[t!]
    \centering
    \begin{subfigure}[t]{0.2\linewidth}
        \centering
        \includegraphics[max height=2.1cm,max width=\linewidth]{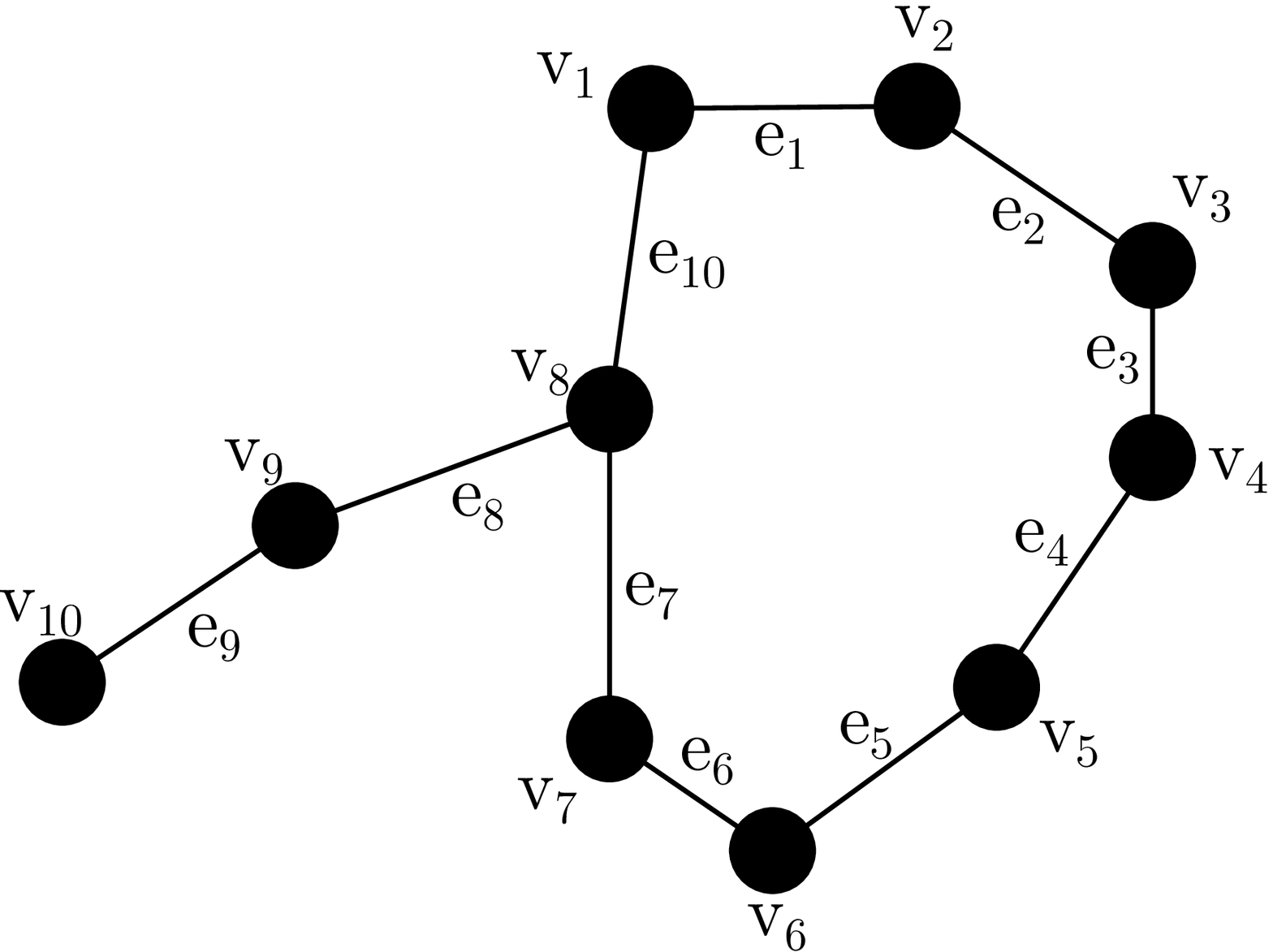}
        \caption{Example graph.} \label{fig:graph_example_a}
    \end{subfigure} \hfill
    \begin{subfigure}[t]{0.19\linewidth}
        \centering
        \includegraphics[max height=2.5cm,max width=\linewidth]{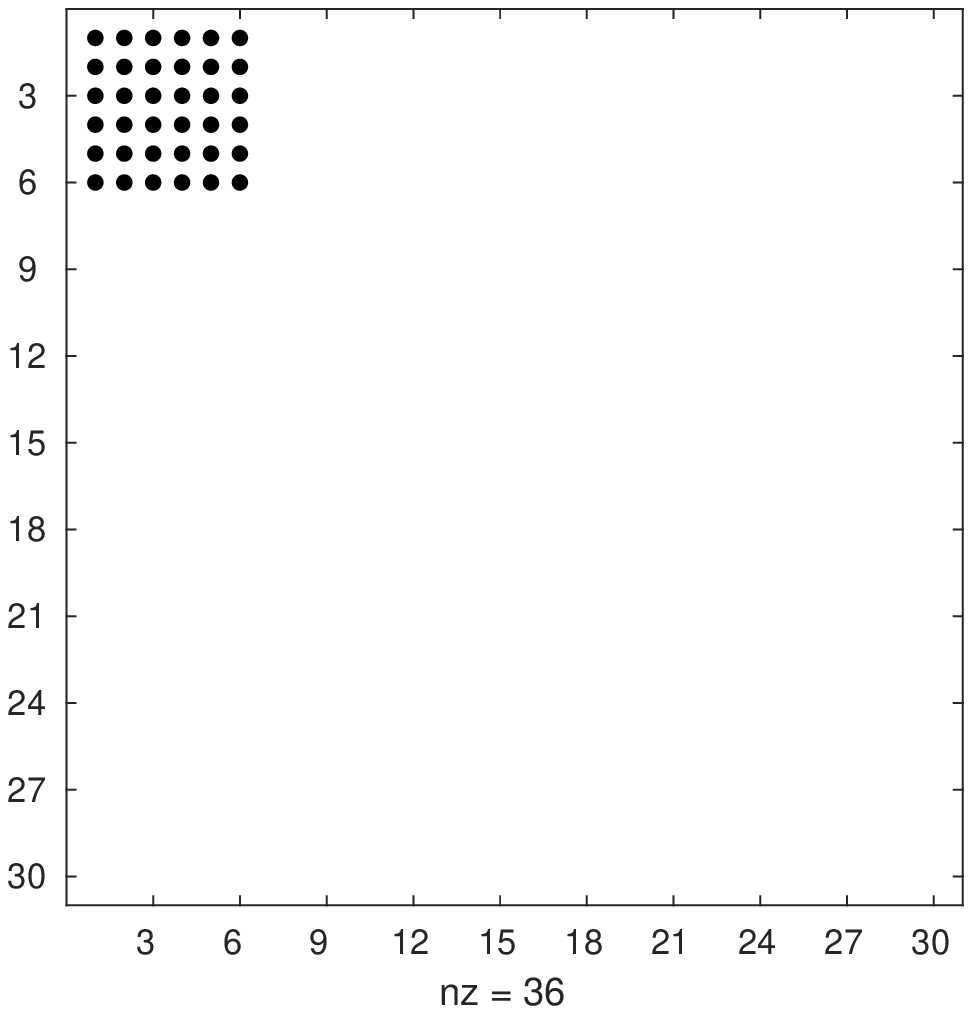}
        \caption{$\boldsymbol{Y}_1$ (odometry edge).} \label{fig:graph_example_b}
    \end{subfigure} \hfill
    \begin{subfigure}[t]{0.19\linewidth}
        \centering
        \includegraphics[max height=2.5cm,max width=\linewidth]{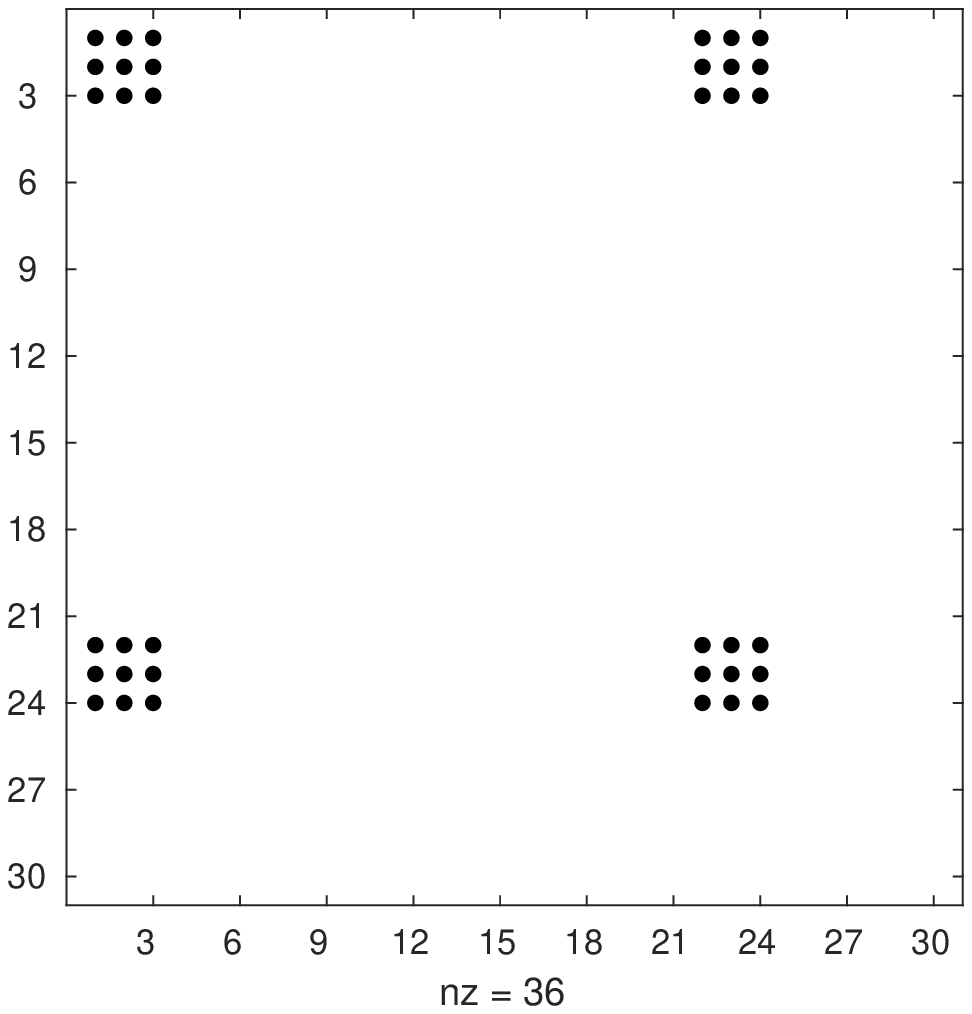}
        \caption{$\boldsymbol{Y}_{10}$ (loop closure).} \label{fig:graph_example_c}
    \end{subfigure} \hfill
    \begin{subfigure}[t]{0.19\linewidth}
        \centering
        \includegraphics[max height=2.5cm,max width=\linewidth]{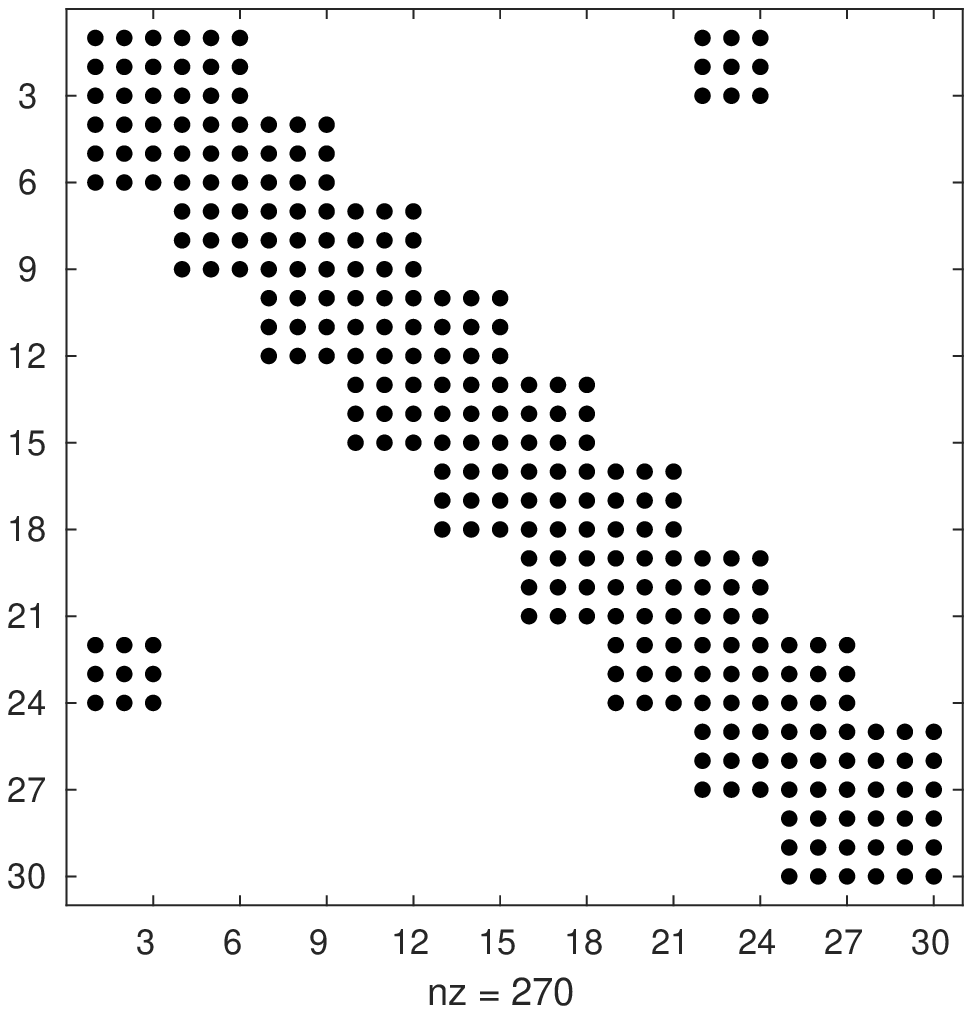}
        \caption{Full FIM, $\boldsymbol{Y}$.} \label{fig:graph_example_d}
    \end{subfigure}
    \begin{subfigure}[t]{0.19\linewidth}
        \centering
        \includegraphics[max height=2.5cm,max width=\linewidth]{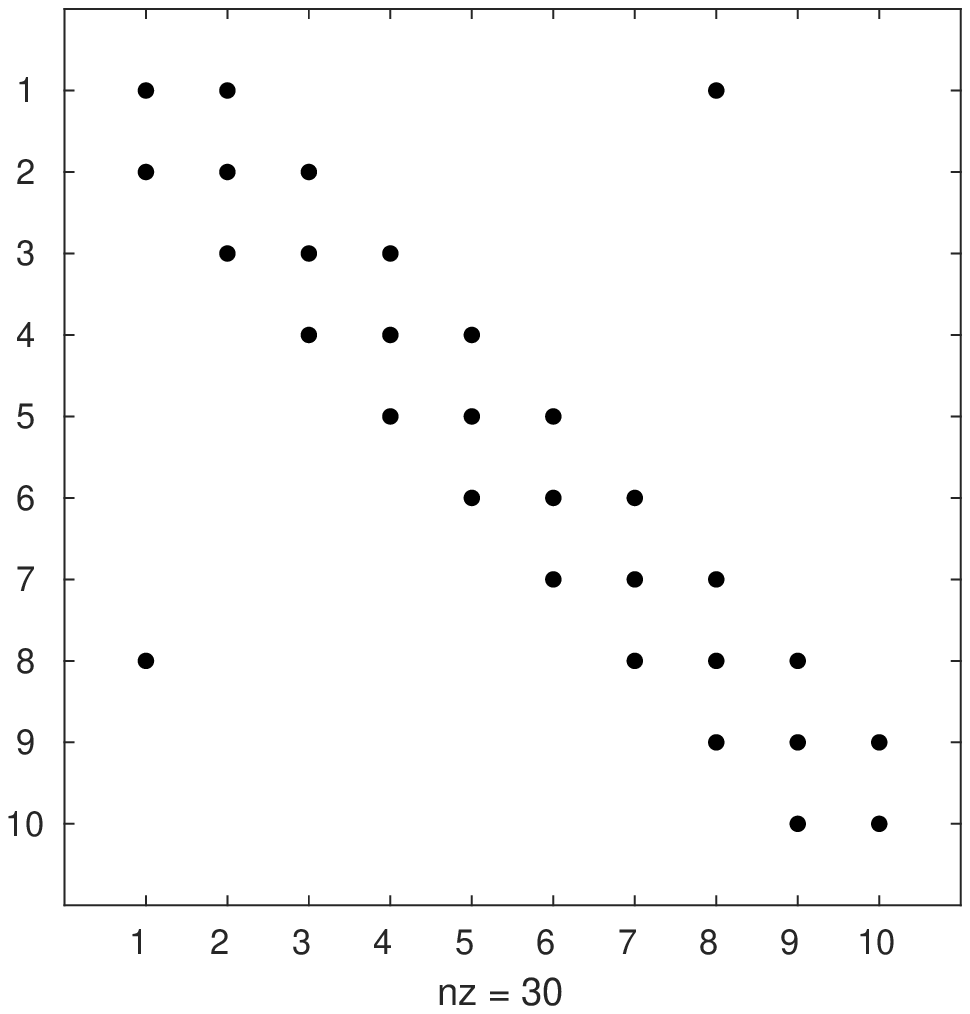}
        \caption{Graph Laplacian, $\boldsymbol{L}$.} \label{fig:graph_example_e}
    \end{subfigure}
    \caption{Example of two of the generators $\boldsymbol{Y}_j$, and $\boldsymbol{Y}$ for an example 2D pose-graph with $n=m=10$ that contains one loop closure. Also, the graph Laplacian. Non-zero matrix elements are depicted with black dots.} \label{fig:graph_example}
\end{figure*}

At this point, we can leverage graph theory and write
${\boldsymbol{\mathcal{I}}_j}=\boldsymbol{q}_j^T\otimes\pmb{\mathbb{I}}_{\ell}$, where $\otimes$ denotes the Kronecker product and $\boldsymbol{q}_j$ is the column vector that identifies the vertices incident upon the $j$-th edge (see Section~\ref{SS:3c}). Then, using the transpose and mixed-product properties of the Kronecker product, the full FIM can be expressed in terms of the pose-graph topology,
\begin{equation}
    \therefore \boldsymbol{Y} = \textstyle\sum\limits_{j=1}^m \boldsymbol{Y}_j = \sum\limits_{j=1}^m \boldsymbol{E}_j \otimes \boldsymbol{\Sigma}_j^{-1} \label{eq:3}
\end{equation}
where the generator $\boldsymbol{Y}_j \in \mathbb{R}^{n\ell\times n\ell}$ is the information matrix of the entire system associated to the $j$-th edge; and $\boldsymbol{E}_j=\boldsymbol{q}_j \boldsymbol{q}_j^T$ are the Laplacian generators, see~\eqref{eq:laplacian}.

The left- and right-multiplication of the covariance matrices of the measurements by ${\boldsymbol{\mathcal{I}}_j}$ confers $\boldsymbol{Y}$ a very special block-sparsity pattern that, in fact, conveys that of the Laplacian of the underlying pose-graph. Figure~\ref{fig:graph_example} contains a pose-graph toy example, for which the information matrix, two of their generators and its Laplacian are shown. Sub-figures~\ref{fig:graph_example_d} and~\ref{fig:graph_example_e} illustrate the similarity between the block-sparsity patterns of $\boldsymbol{Y}$ and the Laplacian matrix, $\boldsymbol{L}$.

Two special cases of equality~\eqref{eq:3} arise, in which it is possible to directly link the full information matrix to the (weighted) Laplacian rather than to its generators.
The first one corresponds to the situation of \mbox{\textbf{constant uncertainty}} through measurements (i.e., a constant covariance matrix or FIM, $\bar{\boldsymbol{\Phi}}$, for all $j$); a common assumption in related literature~\cite{khosoussi14}, although unrealistic except for some exploratory trajectories. Under this hypothesis, by leveraging the associative property of the Kronecker product,~\eqref{eq:3} becomes:
\begin{align}
    \boldsymbol{Y} = \textstyle\sum\limits_{j=1}^m \boldsymbol{E}_j \otimes \bar{\boldsymbol{\Phi}} = \boldsymbol{L} \otimes \bar{\boldsymbol{\Phi}} \qquad  \text{if } \boldsymbol{\Phi}_{j}=\bar{\boldsymbol{\Phi}} \ \forall j \label{eq:4}
\end{align}

The second case considers \textbf{variable uncertainty} along exploration. Since any positive semi-definite matrix can be considered trivially upper-bounded by a diagonal matrix with its largest eigenvalue as diagonal terms, it holds:
\begin{equation}
    \boldsymbol{\Sigma}_j \succeq \lambda^j_1 \ \pmb{\mathbb{I}}_\ell
    \ \ \Leftrightarrow \ \
    \boldsymbol{\Phi}_j \preceq \rho^j_\ell \ \pmb{\mathbb{I}}_\ell = (\lambda^j_1)^{-1} \ \pmb{\mathbb{I}}_\ell \label{eq:bound_eopt}
\end{equation}
with $(\lambda^j_1,\mydots,\lambda^j_\ell)$ the ordered set of eigenvalues of $\boldsymbol{\Sigma}_j$ and $(\rho^j_1,\mydots,\rho^j_\ell)$ that of the FIM $\boldsymbol{\Phi}_j={\boldsymbol{\Sigma}_j}^{-1}$; ranked in increasing order.
Using the previous bound,~\eqref{eq:eopt} and the associative property of the Kronecker product,~\eqref{eq:3} now becomes:
\begin{align}
    \boldsymbol{Y} &\preceq \textstyle\sum\limits_{j=1}^m \left(\|\boldsymbol{\Phi}_j\|_\infty \ \boldsymbol{E}_j \right)\otimes \pmb{\mathbb{I}}_\ell \nonumber\\
    &= \boldsymbol{L}_\gamma \otimes \pmb{\mathbb{I}}_\ell \ \ \ \text{if } \boldsymbol{\Phi}_j\preceq\ \|\boldsymbol{\Phi}_j\|_\infty \ \pmb{\mathbb{I}}_\ell \ \forall j \label{eq:5}
\end{align}
\noindent where $\boldsymbol{L}_\gamma$ is the Laplacian of the pose-graph in which each edge is weighted with $\gamma_j=\|\boldsymbol{\Phi}_j\|_\infty$. As with the weighted Laplacian,~\eqref{eq:5} yields to~\eqref{eq:4} for the case that $\gamma_j = 1 \ \forall j$.

\subsection{On the Spectra} \label{SS:4a}

Consider now $(\bar{\rho}_1,\mydots,\bar{\rho}_\ell)$ to be the ordered set of eigenvalues of $\bar{\boldsymbol{\Phi}}$, and $(0=\mu_1, \mu_2, \mydots,\mu_n)$ that of the Laplacian matrix $\boldsymbol{L}$, again ranked in increasing order. According to the spectral properties of the Kronecker product:
\begin{equation}
   \text{eig}\left(\boldsymbol{L}\otimes\bar{\boldsymbol{\Phi}}\right)= \mu_k\ \bar{\rho}_b, \ \ \begin{array}{ll}
            k=1,\mydots,n \\
            b=1,\mydots,\ell
        \end{array} \label{eq:6}
\end{equation}

Thus, under the assumption of \textbf{constant uncertainty}, optimality criteria applied to $\boldsymbol{Y}$ can be obtained by applying it separately to the reduced Laplacian (i.e., after removing its zero eigenvalue) and $\bar{\boldsymbol{\Phi}}$. For the different $p$-values, it will be:
\begin{align}
    \|\boldsymbol{Y}\|_p &=  \left\{
        \begin{array}{lrr}
            \|\boldsymbol{L}\|_p \ \|\bar{\boldsymbol{\Phi}}\|_p \ & \text{if} &0<|p|<\infty \\
            \|\boldsymbol{L}\|_p \ \|\bar{\boldsymbol{\Phi}}\|_p \ \|\bar{\boldsymbol{\Phi}}\|_p^{-\frac{1}{n}} & \text{if} &p=0
        \end{array} \right. \label{eq:7}
\end{align}
\noindent Leading to the following proportionality relationship,
\begin{equation}
    \therefore \|\boldsymbol{Y}\|_p \propto \|\boldsymbol{L}\|_p \quad \forall p \label{eq:71}
\end{equation}

The above equation allows to shift the traditional approach of applying optimality criteria to $\boldsymbol{Y}$, to a new strategy where they are applied to $\boldsymbol{L}$. Given that new optimality criteria will be functionals of the reduced Laplacian eigenvalues, the spectral graph theory presented in Section~\ref{SS:2d} can be leveraged. Hence, under the assumption that every measurement has the same covariance, optimality criteria of $\boldsymbol{Y}$ (or $\boldsymbol{\Sigma}$) can be expressed in terms of the pose-graph structure as:
\begin{alignat}{2}
    T{\text -} opt(\boldsymbol{Y}) &= A{\text -} opt(\boldsymbol{\Sigma})^{-1} &\propto T{\text -} opt(\boldsymbol{L}) &= \bar{d}\label{eq:t_cte}\\
    D{\text -} opt(\boldsymbol{Y}) &= D{\text -} opt(\boldsymbol{\Sigma})^{-1} &\propto D{\text -} opt(\boldsymbol{L}) &=\left( n\ t({\mathcal{G}})\right)^{\frac{1}{n}} \label{eq:d_cte}\\
    A{\text -} opt(\boldsymbol{Y}) &= T{\text -} opt(\boldsymbol{\Sigma})^{-1} &\propto A{\text -} opt(\boldsymbol{L}) &=n^2\ K({\mathcal{G}})^{-1} \label{eq:a_cte}\\
    E{\text -} opt(\boldsymbol{Y}) &= \tilde{E}{\text -} opt(\boldsymbol{\Sigma})^{-1} &\propto E{\text -} opt(\boldsymbol{L}) &= \alpha({\mathcal{G}}) \label{eq:e_cte}
\end{alignat}
\noindent where $n$ is the number of nodes and $\bar{d}\triangleq2m/n$ the average degree of the pose-graph. Note that~\eqref{eq:d_cte} is consistent with~\cite{khosoussi14} for the particular case they studied in which $\ell=\{2, 3\}$ and $D{\text -} opt$ is defined in a traditional way~\cite{wald43}.

On the other hand, using Weyl's monotonicity theorem~\cite{gohberg78}, the following inequality is satisfied for the case of \textbf{variable uncertainty}, as expressed in~\eqref{eq:5}:
\begin{equation}
    \therefore \|\boldsymbol{Y}\|_p \leq \|\boldsymbol{L}_\gamma\|_p \ \ \text{if} \  \gamma_j=\|\boldsymbol{\Phi}_j\|_\infty \ \forall p \label{eq:8}
\end{equation}
\noindent Thus, we are first weighting the graph edges individually with $\tilde{E}\text{-}opt$ and then computing the desired optimality criteria of the graph. The bound in~\eqref{eq:bound_eopt} and the fact that $\tilde{E}\text{-}opt\geq T\text{-}opt\geq D\text{-}opt\geq A\text{-}opt\geq E\text{-}opt$ makes~\eqref{eq:8} hold for all $p$. If we consider isotropic noise, the bound in~\eqref{eq:bound_eopt} turns into equality and so does~\eqref{eq:8}. However, for non-isotropic (nor diagonal) covariance matrices,~\eqref{eq:8} represents an extremely conservative bound for criteria other than the one used as weight. Interestingly, for $p=\infty$ the bound particularizes to $\tilde{E}{\text -} opt(\boldsymbol{Y})\lesssim\tilde{E}{\text -} opt(\boldsymbol{L}_\gamma)$ because (i) the highest eigenvalue is weakly affected by off-diagonal elements and (ii) absolute values of the off-diagonal terms in SLAM FIMs are generally smaller than those on the main diagonal.

Instead of the loose upper-bound that~\eqref{eq:8} offers in general, an approximation relationship can be obtained by following the strategy of weighting the pose-graph with the same optimality criterion to be estimated:
\begin{equation}
    \therefore \| \boldsymbol{Y}\|_p \approx \|\boldsymbol{L}_\gamma\|_p \ \ \text{if} \  \gamma_j=\|\boldsymbol{\Phi}_j\|_p \ \forall p \label{eq:81}
\end{equation}
The goodness of approximation, in general, will depend on: (i) the number of off-diagonal elements in both $\boldsymbol{\Phi}_j$ (cross-correlations) and $\boldsymbol{Y}$ (loop closures); (ii) the values of the off-diagonal terms in $\boldsymbol{\Phi}_j$, relative to those in the main diagonal; and (iii) how every criterion accounts for them.
Regarding the latter, equality will emerge for $T\text{-}opt$ as
it neglects off-diagonal terms, and $E\text{-}opt(\boldsymbol{L}_\gamma)$ will represent a lower-bound, unlike $\tilde{E}{\text -} opt(\boldsymbol{L}_\gamma)$, although more affected by off-diagonal terms. Finally, particularizing~\eqref{eq:81} for the different $p$-values,
\begin{align}
    \tilde{E}{\text -} opt(\boldsymbol{Y})&\lesssim\tilde{E}{\text -} opt(\boldsymbol{L}_\gamma)\label{eq:te_vble}\\
    T{\text -} opt(\boldsymbol{Y}) &= T{\text -} opt(\boldsymbol{L}_\gamma) = \bar{d}\ \bar{\gamma}\label{eq:t_vble}\\
    D{\text -} opt(\boldsymbol{Y}) &\approx D{\text -} opt(\boldsymbol{L}_\gamma)=\left( n\ t({\mathcal{G}}_\gamma)\right)^{\frac{1}{n}}\label{eq:dopt_graph}\\
    E{\text -} opt(\boldsymbol{Y}) &\gtrsim E{\text -} opt(\boldsymbol{L}_\gamma)= \alpha({\mathcal{G}}_\gamma)\label{eq:e_vble}
\end{align}
\noindent where ${\mathcal{G}}_\gamma$ now denotes the pose-graph weighted with the same criterion to be computed, $\bar{\gamma}$ its average weight and $\boldsymbol{L}_\gamma$ its weighted Laplacian.
$A{\text -} opt$ was not presented since the complexity of
the weighted Kirchhoff index makes its use worthless.
Also, computation of the Laplacian determinant to evaluate the number of spanning trees quickly becomes intractable for large graphs, as well as it generates low precision for small values. The logarithmic determinant avoids under/overflow and allows to compute~\eqref{eq:dopt_graph} efficiently via:
\begin{equation}
    D{\text -} opt(\boldsymbol{Y}) \approx n^{\frac{1}{n}} \ \exp{\left\{\log(t({\mathcal{G}}_\gamma))/n\right\}}
\end{equation}

\section{Experimental Validation} \label{S:5}

In this section, we conduct several experiments to prove the theoretical relationships in~\eqref{eq:t_cte}-\eqref{eq:e_cte} and~\eqref{eq:te_vble}-\eqref{eq:e_vble} hold. Pose-graph datasets from~\cite{carlone14} and~\cite{carlone15}, which are publicly available,
have been used for 2D and 3D experiments, respectively.
To compare traditional and graph-based approaches, we simulate the construction of the pose-graph as if the robot were performing active SLAM. That is, at each time step, a new node and its corresponding constraints are added to the graph and optimality criteria is computed using both $\mathbf{Y}$ and $\mathbf{L}$. Experiments have been performed on an Intel Core i9 CPU.

\subsection{Constant Uncertainty Case} \label{SS:5a}

First, we used a reduced trajectory of the FRH 2D dataset that contains the path before the first loop closure occurs (purely exploratory). We assigned to every edge the same following FIM, which is non-isotropic and in which the translational variances are correlated:
\begin{align}
    \boldsymbol{\Phi}_j = \bar{\boldsymbol{\Phi}} &=
    \scalebox{.9}{%
        \renewcommand{\arraystretch}{.8}%
        $\begin{pmatrix} 11.11 &-3 &0 \\ -3 &6.25 &0 \\ 0 &0 &250\end{pmatrix}$
      }
      \ \forall j\label{eq:ct_cov_exp}
\end{align}

\begin{figure}[!t]
    \centering
    \vspace*{4pt}
    \begin{subfigure}[b]{0.48\linewidth}
        \centering
        \includegraphics[max width=\linewidth, max height=3.1cm]{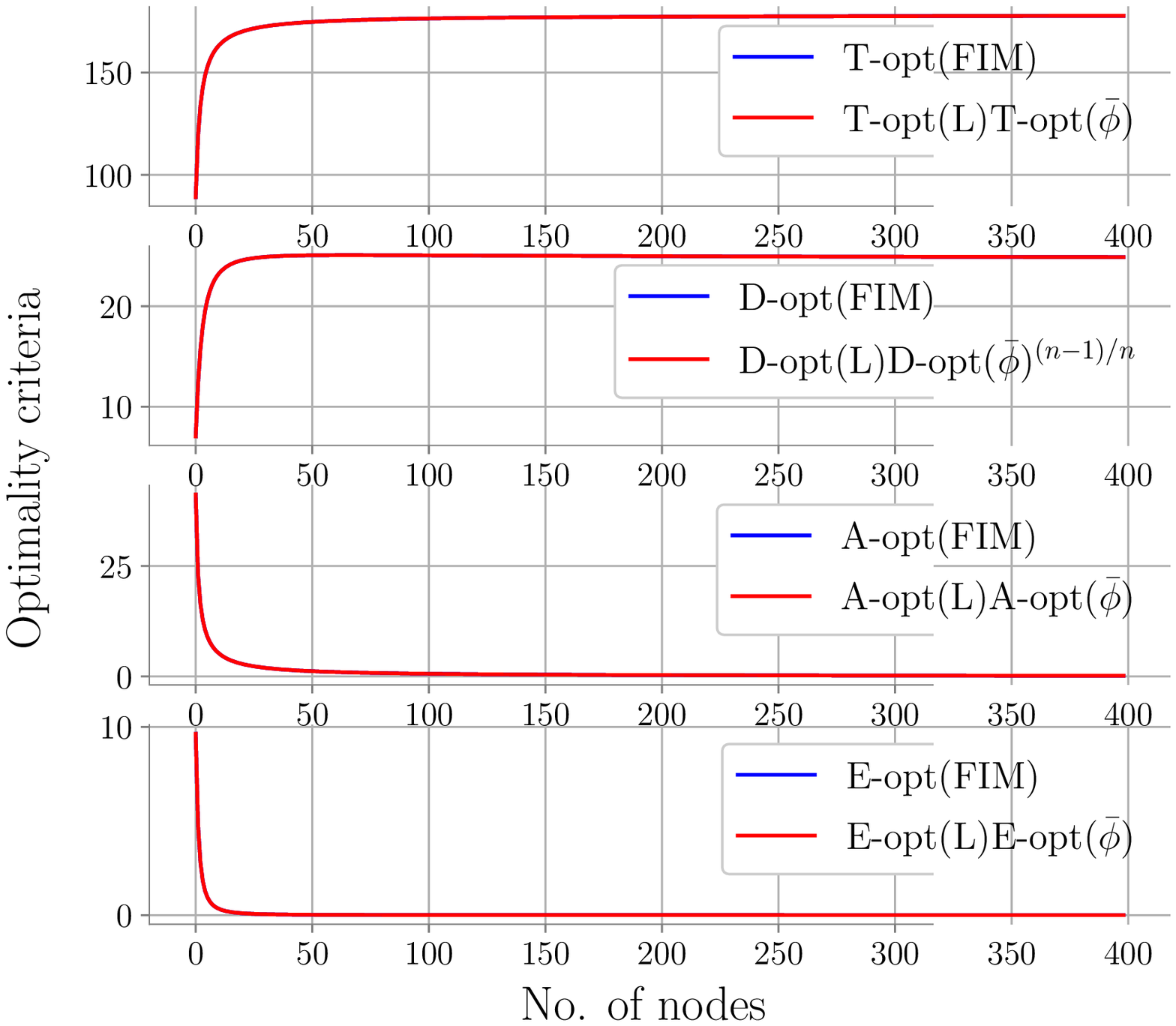}
        \caption{Optimality criteria.} \label{fig:FRH_comparison_fim}
    \end{subfigure} \hfill
    \begin{subfigure}[b]{0.5\linewidth}
        \centering
        \includegraphics[max width=\linewidth, max height=3.1cm]{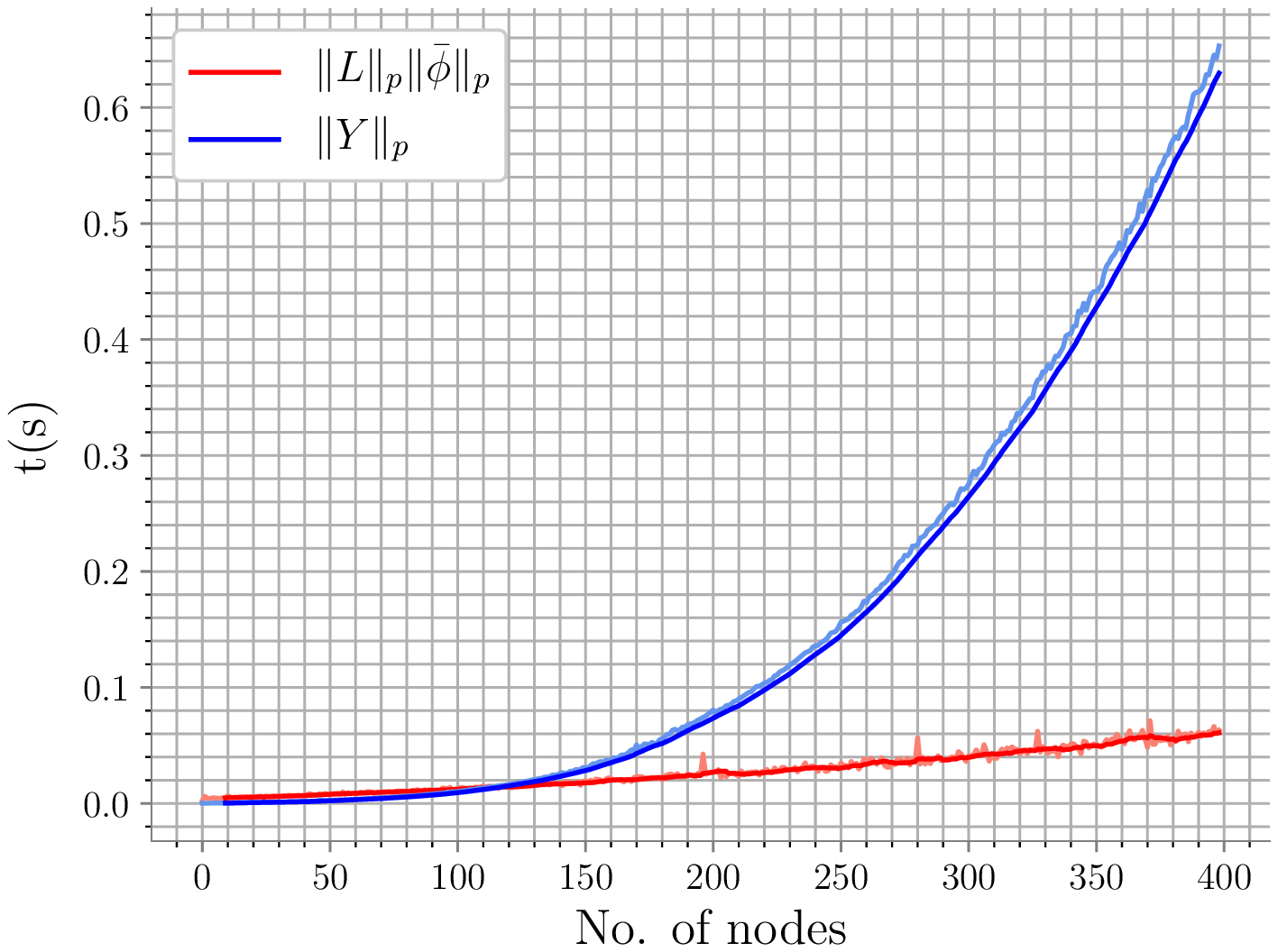}
        \caption{Time ($s$).} \label{fig:FRH_time_fim}
    \end{subfigure}
    \caption{Optimality criteria of the full FIM (blue) and the Laplacian (red), and the time required per step to compute them in the reduced FRH sequence with constant uncertainty.}
\end{figure}

Figure~\ref{fig:FRH_comparison_fim} shows the computed $T\text{-},D\text{-},A\text{-}$ and $E\text{-}opt$ using the estimation-theoretic (blue) and graphical (red) facets of the problem. All curves overlap, proving that the relationships in~\eqref{eq:t_cte}-\eqref{eq:e_cte} hold and, moreover, that the proportionality constants derived in~\eqref{eq:7} are consistent ---although they are not necessary for active SLAM. Time consumed per step by both approaches appears in Figure~\ref{fig:FRH_time_fim}, clearly showing the advantage of computing $\|\boldsymbol{L}\|_p \|\bar{\boldsymbol{\Phi}}\|_p$ (red) over $\|\boldsymbol{Y}\|_p$ (blue) as the size of the pose-graph grows. Studying computational complexity, optimality criteria on $\boldsymbol{L}$ requires $\mathcal{O}(n^3)+\mathcal{O}(\ell^3)$ while on $\boldsymbol{Y}$ requires $\mathcal{O}(\ell^3n^3)$, omitting lower order terms and being $\mathcal{O}(\cdot)$ a lower-bound. Also, as the FIM dimension grows, building $\mathbf{Y}$ requires way far more resources than creating ${\mathcal{G}}$.

\subsection{Variable Uncertainty Case} \label{SS:4b}

\begin{figure*}[t!]
    \centering
    \begin{subfigure}[b]{0.24\linewidth}
        \centering
        \includegraphics[max width=\linewidth, max height=8cm]{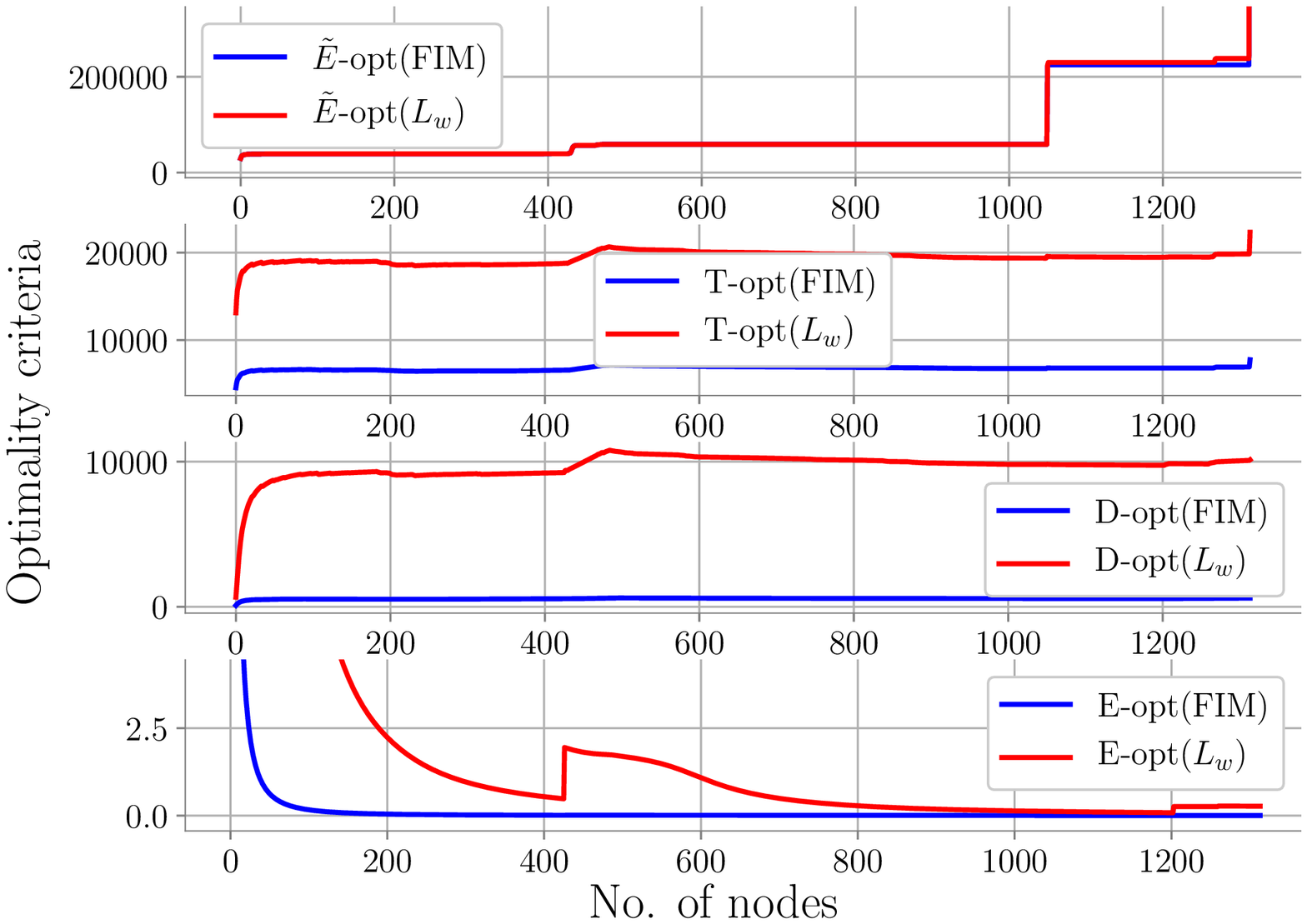}
        \caption{FRH, $\gamma_j=\|\boldsymbol{\Phi}_j\|_\infty$.}
        \label{fig:FRH_comparison_FIM_var_full_1}
    \end{subfigure} \hfill
    \begin{subfigure}[b]{0.24\linewidth}
        \centering
        \includegraphics[max width=\linewidth, max height=8cm]{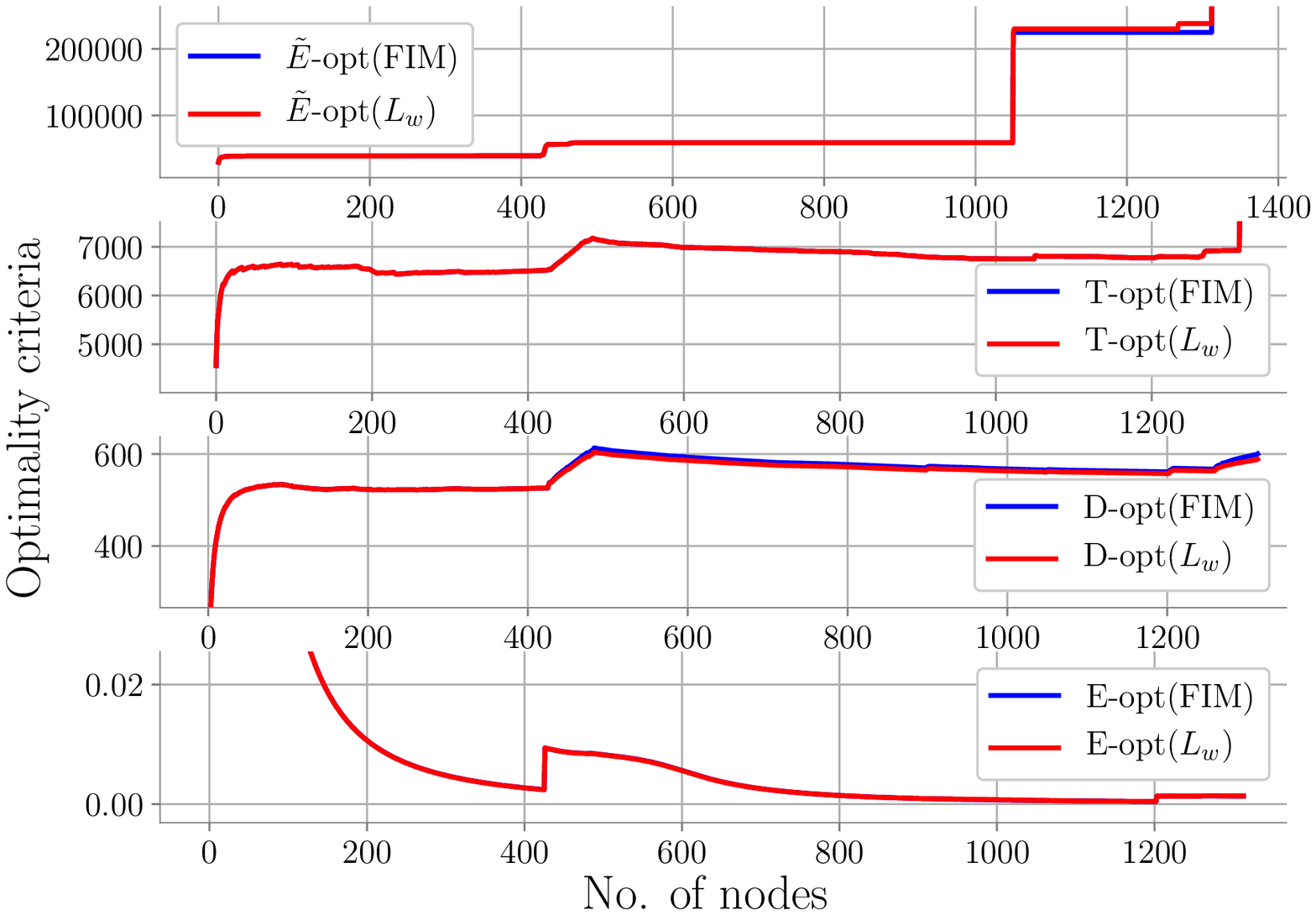}
        \caption{FRH, $\gamma_j=\|\boldsymbol{\Phi}_j\|_p$.}
        \label{fig:FRH_comparison_FIM_var_full_2}
    \end{subfigure} \hfill
    \begin{subfigure}[b]{0.24\linewidth}
        \centering
        \includegraphics[max width=\linewidth, max height=8cm]{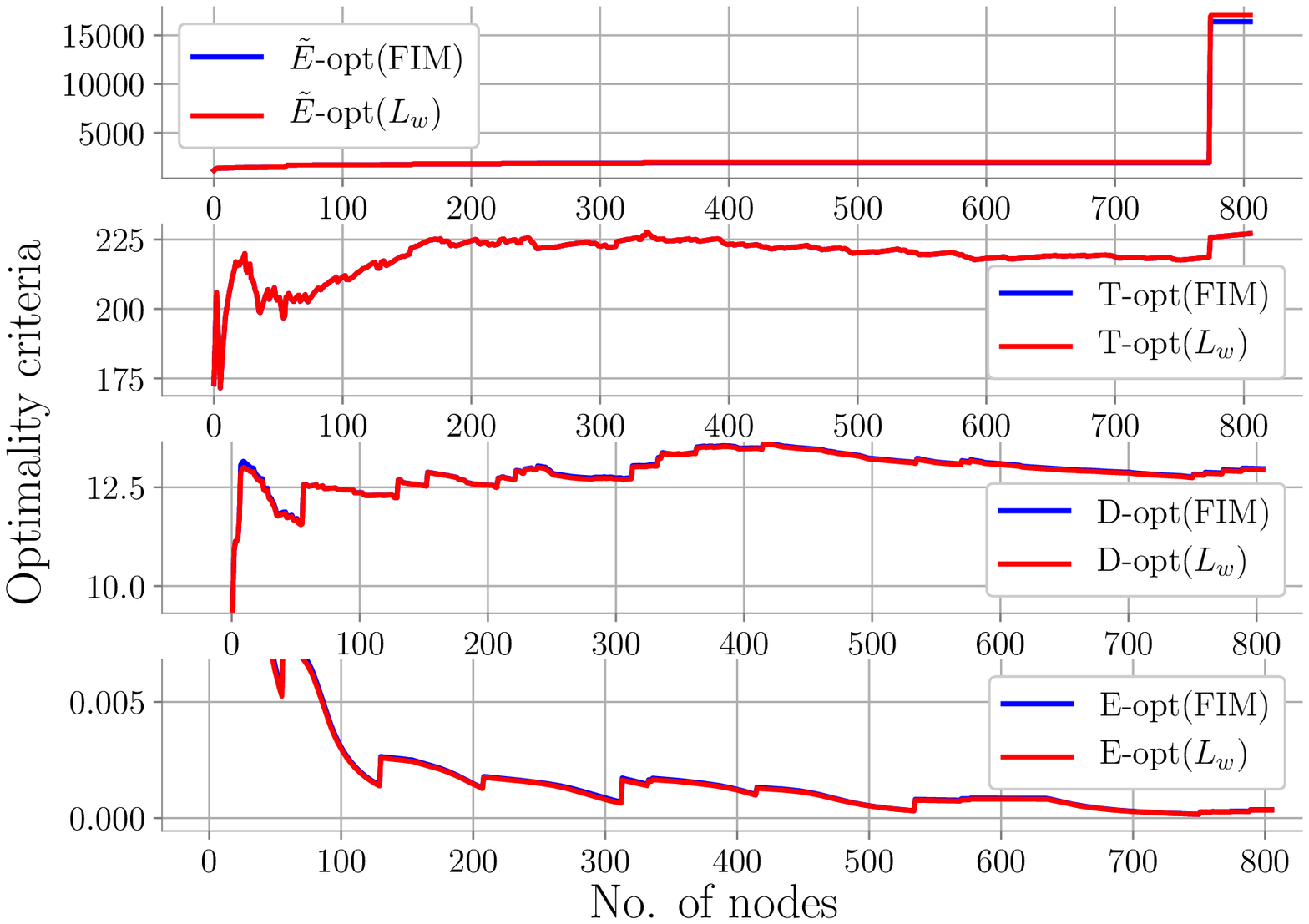}
        \caption{MIT,  $\gamma_j=\|\boldsymbol{\Phi}_j\|_p$.}
        \label{fig:MITb_comparison_FIM_var_full}
    \end{subfigure} \hfill
    \begin{subfigure}[b]{0.24\linewidth}
        \centering
        \includegraphics[max width=\linewidth, max height=8cm]{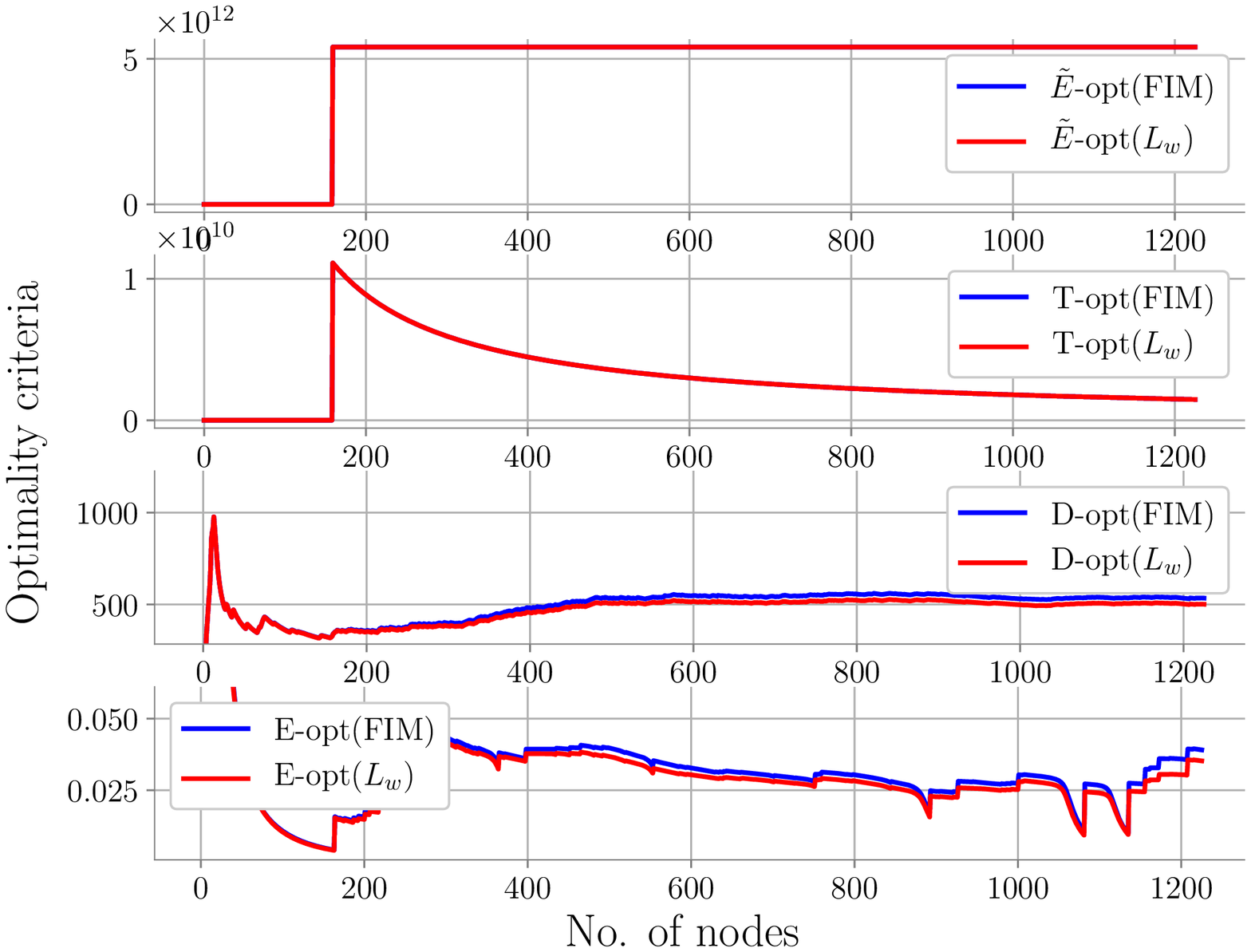}
        \caption{INTEL,  $\gamma_j=\|\boldsymbol{\Phi}_j\|_p$.}
        \label{fig:INTEL_comparison_FIM_var_full}
    \end{subfigure}
    \caption{Optimality criteria of the full FIM (blue) and the Laplacian (red) weighted with $\|\boldsymbol{\Phi}_j\|_\infty$ (a), and $\|\boldsymbol{\Phi}_j\|_p$ (b-d).} \label{fig:comparison_FIM_var_full}
\end{figure*}

Consider now the case in which edges' FIMs are no longer constant. According to~\eqref{eq:8}, one first needs to construct a graph weighted with $\gamma_j=\|\boldsymbol{\Phi}_j\|_\infty$. In this case, the whole trajectory and the original FIMs contained in FRH dataset have been used (which sparsity structure is similar to~\eqref{eq:ct_cov_exp}). Figure~\ref{fig:FRH_comparison_FIM_var_full_1} contains the resulting $\tilde{E}\text{-},T\text{-},D\text{-}$ and $E\text{-}opt$ for the full information matrix (blue) and the weighted Laplacian (red). The selected bound indeed limits $\|\boldsymbol{Y}\|_p \ \forall p$, though it is an extremely conservative one for $p$ other than $\infty$, for which~\eqref{eq:te_vble} holds during the entire sequence.

\begin{figure}[!t]
    \centering
    \begin{subfigure}[b]{0.48\linewidth}
        \centering
        \includegraphics[max width=\linewidth, max height=6.4cm]{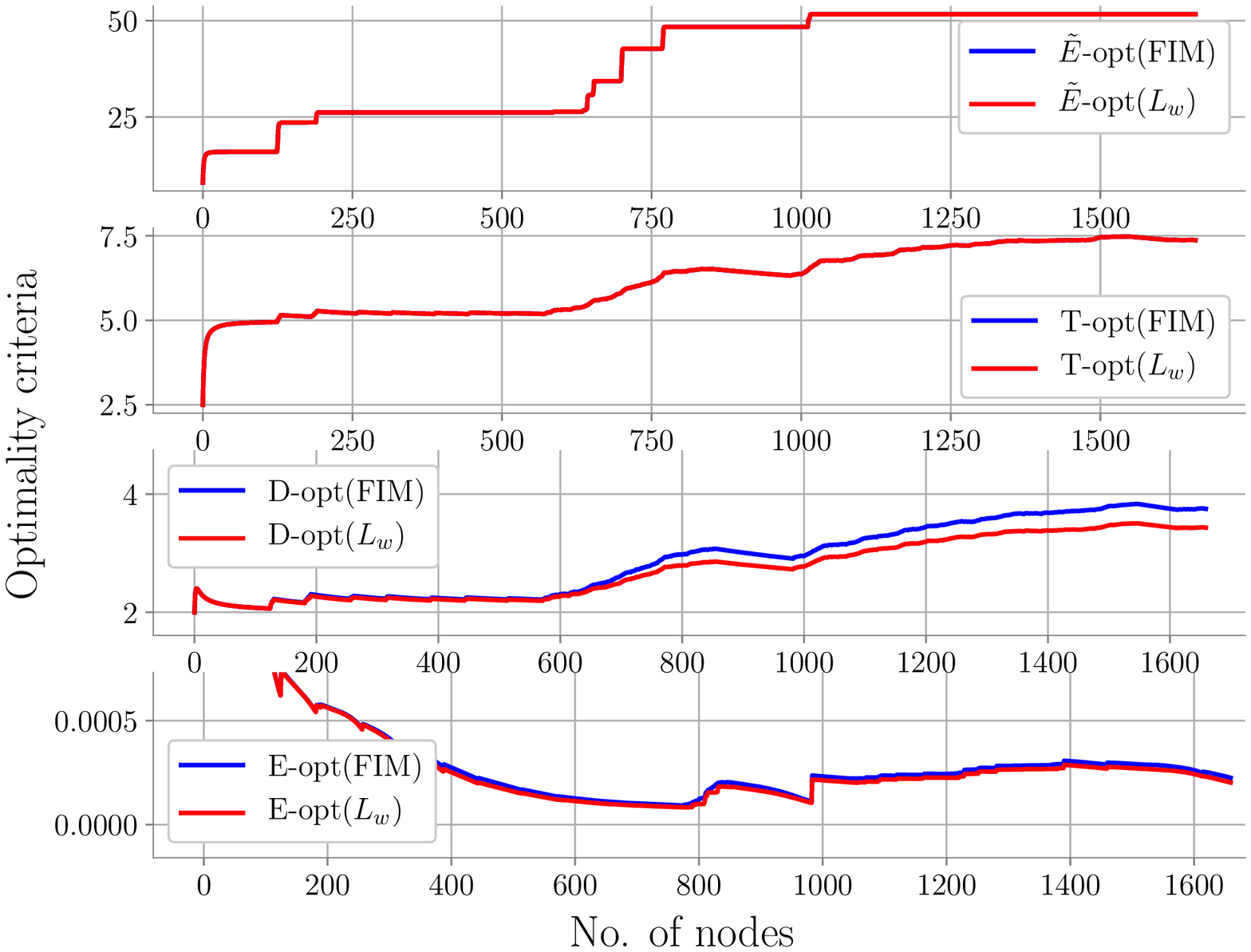}
        \caption{Optimality criteria.} \label{fig:garage_comparison_FIM}
    \end{subfigure} \hfill
    \begin{subfigure}[b]{0.5\linewidth}
        \centering
        \includegraphics[max width=\linewidth, max height=6.4cm]{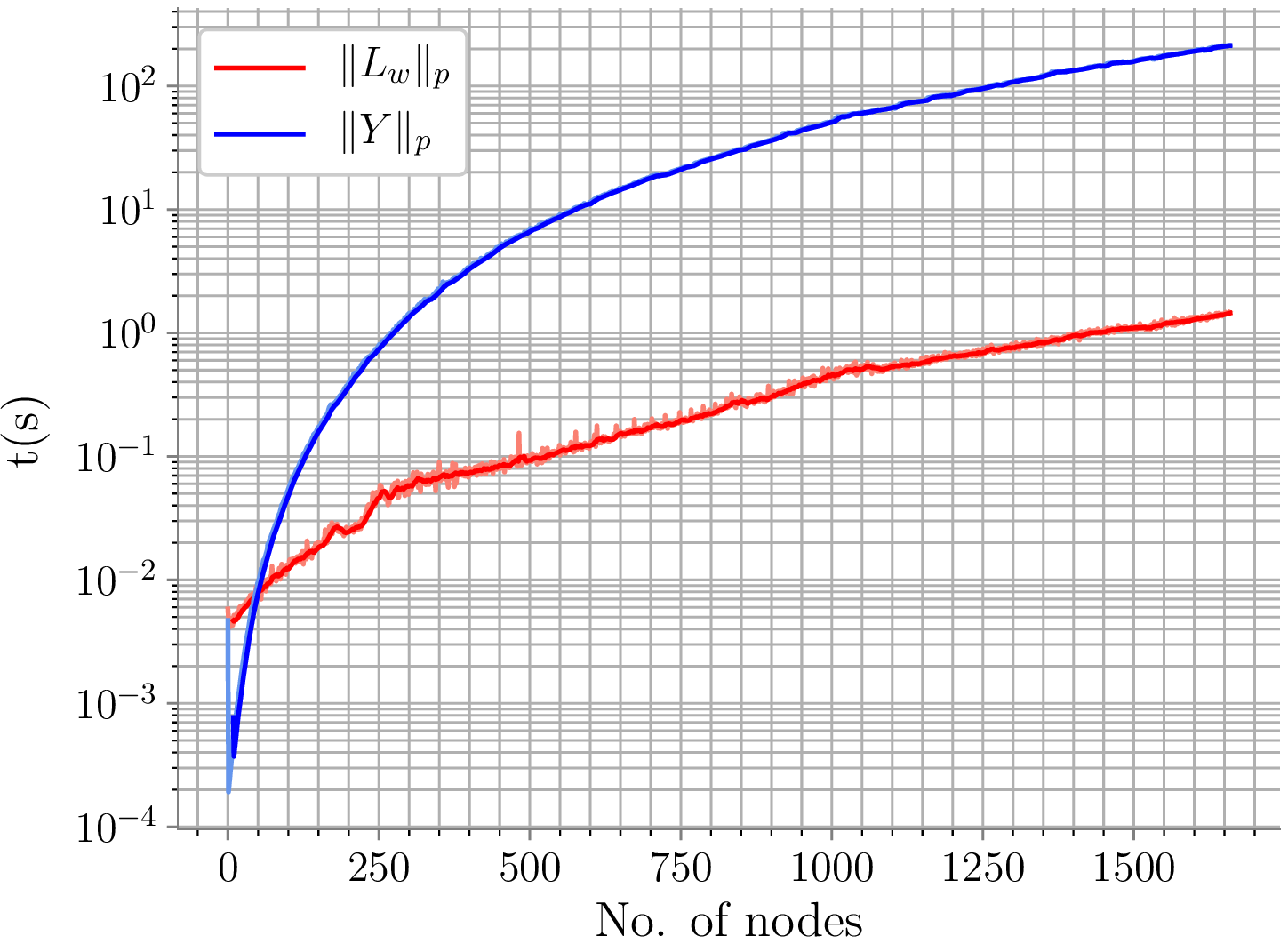}
        \caption{Time ($s$).} \label{fig:garage_comparison_FIM_time}
    \end{subfigure}
    \caption{Optimality criteria of the full FIM (blue) and the Laplacian (red) in the Garage dataset. Also, the time required per step to compute them.}
\end{figure}

Figure~\ref{fig:FRH_comparison_FIM_var_full_2} shows the results in the same dataset using the approximation in~\eqref{eq:81} instead. Blue and red curves are now much closer to each other; overlapping for \textit{T-opt}, and also during certain parts of the trajectory for the other criteria. Besides, the trend of the two curves is the same (i.e., either both increase or decrease); a property that holds in all studied datasets and key for active SLAM ---otherwise we could be wrongly detecting an information gain/loss. 
Further experiments have been carried out using the MIT, Intel, and 3D Garage datasets; to prove the proposed relationships hold and that are not dependent on the dimension of the estimation vector. Analogous results to those seen in FRH dataset have been obtained, and they are presented in Figures~\ref{fig:MITb_comparison_FIM_var_full},~\ref{fig:INTEL_comparison_FIM_var_full} and~\ref{fig:garage_comparison_FIM}, respectively.
Figure~\ref{fig:garage_comparison_FIM_time} shows the time difference in the 3D dataset. Behavior is equivalent to Figure \ref{fig:FRH_time_fim}, although now appears in a log-linear plot and the difference grows up to $10^2 s$ in the end of the sequence.

\renewcommand{\arraystretch}{1.1}
\begin{table*}[t!]
    \footnotesize
    \centering
    \vspace*{4pt}
    \begin{tabular}{l||r|r|c||c|c|c|c||r|r|c}
        \multirow{2}{*}{\textbf{Dataset} }& \multicolumn{3}{c||}{} &\multicolumn{4}{c||}{\textbf{Approximation Error}}&
        \multicolumn{3}{c}{\textbf{Time (min)}}\\
        &\textbf{n} & \textbf{m} & \textbf{d} & \textbf{$\boldsymbol{\Delta}\tilde{\textbf{E}}$-opt} & $\boldsymbol{\Delta}$\textbf{T-opt} & $\boldsymbol{\Delta}$\textbf{D-opt} & \textbf{$\boldsymbol{\Delta}\textbf{E}$-opt} &
        \textbf{t($\|\boldsymbol{Y}\|_p$)}      &
        \textbf{t($\|\boldsymbol{L}_\gamma\|_p$)} & \textbf{$\boldsymbol{\Delta}$t} \\ \hline
        MIT & 807 & 827 & 2.1 & 2.76\% & $\sim$0\% & 0.16\% & 3.53\% & 13.75  & 1.96 & 85.7\%\\ \hline
        FR079 & 989 & 1217 & 2.4 & 0.43\% & $\sim$0\% & 7.15\% & 5.33\% & 26.50 & 3.07 & 88.4\%\\ \hline
        CSAIL & 1045 & 1171 & 2.2 & 0.11\% & $\sim$0\% & 1.68\% & 1.05\% & 33.90 & 2.73 & 91.9\% \\ \hline
        INTEL & 1227 & 1481 & 2.4 & $\sim$0\% & $\sim$0\% & 5.85\% & 7.19\% & 63.29 & 7.46 & 88.2\%\\ \hline
        FRH & 1316 & 1485 & 2.2 & $\sim$0\% & $\sim$0\% & 0.76\% & 0.49\% & 121.77 & 5.42 & 95.6\% \\ \hline\hline
        MH01 (monocular) & 376 & 544 & 2.9 & $\sim$0\% & $\sim$0\% & 0.74\% & 1.49\% & 4.22 & 0.56 & 86.6\% \\ \hline
        V101 (monocular) & 264 & 415 & 3.1 & $\sim$0\% & $\sim$0\% & 0.13\% &3.27\% & 1.01 & 0.14 & 84.9\% \\ \hline
        V101-103 (stereo-multi)& 322 & 369 & 2.3 & $\sim$0\% & $\sim$0\% & 0.40\% & 3.65\% & 2.21 & 0.30 & 86.1\% \\ \hline
        V201 (stereo-inertial)& 337 & 598 & 3.5 & $\sim$0\% & $\sim$0\% & 0.79\% & 0.05\% & 2.65 & 0.16 & 94.0\% \\ \hline
        Garage & 1661 & 2615 & 3.1 & 0.01\% & $\sim$0\% & 6.21\% &7.55\% & 1549.9 & 11.70 & 99.2\% \\ \hline\hline
        \textbf{Mean} & - & - & - & \textbf{0.34\%} & \textbf{$\sim$0\%} & \textbf{2.38\%} & \textbf{3.36\%} & -& - & \textbf{90.1\%}
    \end{tabular}
    \caption{Percentage error (median) in estimation of optimality criteria using the graph Laplacian instead of the full FIM. Also, the accumulated time required to compute both approaches (in minutes) and the time reduction achieved.}
    \label{tab:errors}
\end{table*}

Since it is hard to visually capture the exact difference between the curves presented given their similarity, Table~\ref{tab:errors} contains the median percentage errors when using our approximations instead of the expensive computations over $\boldsymbol{Y}$. It contains the results for all the previously mentioned datasets, but also for FR079 and CSAIL (2D). Furthermore, we analyzed some EuRoC sequences~\cite{burri16}, extracting their pose-graphs with ORB-SLAM3~\cite{campos21}.
Quantitative results denote that differences between computing $\|\boldsymbol{L}_\gamma\|_p$ and $\|\boldsymbol{Y}\|_p$ following~\eqref{eq:81} are indeed very low, and even indistinguishable in some cases. $T\text{-}opt$ was perfectly computed in all cases, showing numerical errors only and proving equality in~\eqref{eq:t_vble} holds.
Error in the case of $\tilde{E}\text{-}opt$ is akin ($0.3\%$ on average and being nearly zero in most datasets) since it is insensitive to elements outside the main diagonal when they are lower than those on it. However, for MIT, FR079 and CSAIL the error is not negligible. We attribute it to certain isolated measurements in these datasets with FIMs several orders of magnitude larger than the rest, and with extremely high off-diagonal elements (also orders of magnitude); perhaps due to incorrect behaviors of the SLAM algorithm. In any case, when these measurements are corrected, or when we use a modern SLAM system as in the EuRoC sequences, the errors drop to zero. Besides, the upper-bound in~\eqref{eq:te_vble} is satisfied whenever the error is non-zero.
Unlike the above two, the smallest eigenvalue is much more sensitive to off-diagonal terms (both cross-correlations and loop closures), even when they are small. The estimation error of $E\text{-}opt$ is the highest ($3.4\%$), also due to the numerical complexity of precisely computing the smallest eigenvalue of high-dimensional matrices.
$D\text{-}opt$ subtly inherits the same behavior since it takes into account all eigenvalues, showing an average approximation error of $2.4\%$. Nevertheless, in most cases it remained below $0.8\%$.
A deeper analysis on the eigenvalue variation demonstrates that $\Delta\tilde{E}\text{-}opt$, $\Delta D\text{-}opt$ and $\Delta E\text{-}opt$ increase slightly as the amount of loop closures does. When the edge FIMs become denser and their off-diagonal terms dominate, particularly $\Delta E\text{-}opt$ is
affected.

Also, Table~\ref{tab:errors} contains the total time required to compute modern optimality criteria with both approaches and the reduction achieved using the Laplacian. These computations require, in one case: building the full FIM and computing its optimality criteria; and in the other case:
building 4 different weighted graphs (one for each criterion), analyzing their connectivity and computing optimality criteria equivalences. To make a fair comparison, both methods are evaluated under the same conditions, and eigenvalue computations leverage fast decomposition techniques. On average, computing optimality criteria of the weighted Laplacian required just $10\%$ of the time that traditional calculations over the FIM did.

\section{Conclusions}\label{S:6}

In this paper, we have shown that quantifying uncertainty in active graph-SLAM formulated over
$SE(n)$ can be efficiently done by analyzing the topology of the underlying pose-graph. We have proposed and validated relationships between modern optimality criteria and graph connectivity indices; showing that equivalent results can be obtained in a fraction of the time. On average, approximations with $2\%$ error can be computed in just $10\%$ of the time. Regardless of the estimation error, the same trend is always maintained, proving
that minimum uncertainty (i.e., optimal) actions can be
found by exploiting the graphical structure of the problem.

As future work, we aim to use the proposed method to perform online (multi-robot) active visual SLAM. Also, the use of full graphs instead of only pose-graphs is to be studied.

\section*{Acknowledgments}
This work was partially supported by MINECO project PID2019‐108398GB‐I00 and DGA\_FSE T45\_20R.


\bibliographystyle{IEEEtran}
\bibliography{root}

\vfill

\end{document}